\documentclass[11pt]{article}

\PassOptionsToPackage{hyperfootnotes=false}{hyperref}
\usepackage{acl}
\usepackage{times}
\usepackage{latexsym}
\usepackage[T1]{fontenc}
\usepackage[utf8]{inputenc}
\usepackage{microtype}
\usepackage{inconsolata}
\usepackage{graphicx}
\usepackage{amsmath}
\usepackage{amssymb}
\usepackage{overpic}
\usepackage{multirow}
\usepackage{array}
\usepackage{booktabs}
\usepackage{xcolor}
\usepackage{colortbl}
\usepackage{CJKutf8}
\usepackage{subcaption}
\usepackage[most]{tcolorbox}
\usepackage[capitalize]{cleveref}

\newcommand{\down}[1]{{\scriptsize\textcolor{red!80!black}{$\downarrow$#1\%}}}

\definecolor{promptblue}{RGB}{115, 192, 222}
\definecolor{promptgreen}{RGB}{117, 186, 117}
\definecolor{promptyellow}{RGB}{250, 200, 88}
\newtcblisting{promptlisting}[3][Prompt]{
  colback=black!5!white,
  arc=5pt,
  boxrule=0.5pt,
  fonttitle=\bfseries,
  title={#1},
  colframe=#2,
  breakable,
  listing only,
  listing file=#3,
  listing options={
    basicstyle=\ttfamily\scriptsize,
    breaklines=true,
    columns=fullflexible,
    keepspaces=true,
  },
}

\title{AppSim-Bench: Bridging Real-world Apps and Reproducible Evaluation for Mobile GUI Agents}

\author{
  \textbf{Jintian Feng\textsuperscript{1,2}}\thanks{Work done as a research intern at Acrab AI.},
  \textbf{Long Chen\textsuperscript{3}},
  \textbf{Xiao Yu\textsuperscript{1,2}},
  \textbf{Jiayi Dai\textsuperscript{1,2}},
  \textbf{Chenglong Liu\textsuperscript{2}},\\
  \textbf{Haoru Wang\textsuperscript{2}},
  \textbf{Zizhen Xue\textsuperscript{3}},
  \textbf{Yuxuan Shi\textsuperscript{1,2}}\thanks{Yuxuan Shi is the corresponding author.},
  \textbf{Ziyang Wang\textsuperscript{3}},
  \textbf{Yichen Gong\textsuperscript{3}}
  \\
  \\
  \textsuperscript{1}Laboratory for Artificial Intelligence and New Forms of Education, Central China Normal University\\
  \textsuperscript{2}Faculty of Artificial Intelligence in Education, Central China Normal University\\
  \textsuperscript{3}Agentic Labs, Acrab AI\\
  \small{\texttt{fjt2018@mails.ccnu.edu.cn}}
}

\begin{document}
\maketitle

\begin{abstract}
Mobile GUI agents can execute tasks from natural-language instructions, but their evaluation remains difficult to make both realistic and reproducible. Existing benchmarks typically trade off these goals: simplified apps lack real-world mobile complexity, whereas live commercial apps introduce uncontrolled variation from recommendations, advertisements, accounts, and changing content.
We propose AppSim-Bench, which addresses this trade-off through controllable simulated apps that preserve task-relevant interaction logic while supporting deterministic evaluation. Built through a coding-agent-assisted and human-verified workflow, it contains 557 tasks across 17 high-frequency Chinese and English apps. Its controllable backend data and outcome-based verification remove major sources of environmental stochasticity, enabling reproducible cross-model comparison.
Evaluating 19 GUI agents, spanning general-purpose and GUI-specialized systems, we find that autonomous mobile execution remains far from solved. The best model completes only 50.27\% of tasks, and 28.55\% of tasks are not solved by any agent. Further analysis shows that failures concentrate in longer workflows, numerical reasoning tasks, and inefficient trajectories marked by high action overhead and budget exhaustion. Our project is available at \url{https://github.com/Acrab-Agentic-Labs/AppSim}.
\end{abstract}

\begin{figure*}[!h]
   \begin{overpic}[width=\textwidth]{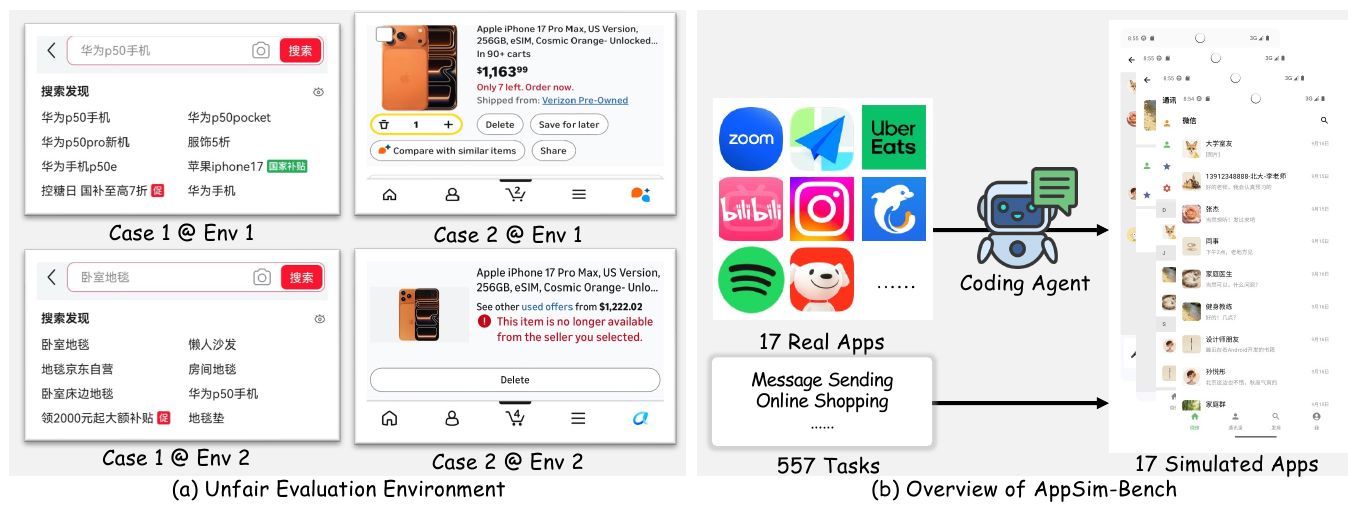} \small
   \end{overpic}
    \caption{
    (a) Two cases showing inconsistent app states across time or devices. In Case 1, \textbf{search suggestions} differ across environments and may affect subsequent agent actions. In Case 2, \textbf{shopping-cart availability} changes across environments, making fair comparison difficult when the task depends on this state.
    (b) Overview of AppSim-Bench, which constructs 17 simulated apps covering 557 tasks.
    }
   \label{fig_Device_Compare}
\end{figure*}

\section{Introduction} \label{sec:intro}

With the rapid development of Vision-Language Models (VLMs), Mobile GUI Agents are emerging as an important research direction in Human-Computer Interaction (HCI)\cite{learnact}. These agents\cite{wang2025opencuaopenfoundationscomputeruse, AgentS2, mobileagent_v3, uitars_1, AutoWebGLM, AppAgent} can understand natural language instructions and autonomously complete multi-step tasks on mobile devices. However, how to systematically and reproducibly evaluate their true capabilities remains a critical, unsolved problem\cite{liu2025llmpoweredguiagentsphone}.

Early evaluation methods often judge task completion by comparing the agent's execution path to a human-annotated ``golden'' path\cite{agent_cpm_CAGUI,rawles2023androidwildlargescaledataset}. 
However, such action-based approaches struggle with the common ``multi-path problem'' in mobile applications, making it difficult to measure the agent's true task-completion ability\cite{learnact, liu2025llmpoweredguiagentsphone, nguyen2025guiagentssurvey}.
For example, in a shopping task, an agent may complete the purchase through a valid but unannotated path.
A trace-based evaluator would still mark this attempt as a failure, exposing a basic limitation of static benchmarks.

To overcome these limitations, dynamic benchmarks are becoming the mainstream approach. 
Represented by AndroidWorld\cite{android_world}, these benchmarks provide agents with a realistic and interactive environment to complete tasks autonomously. 
Success is judged by monitoring changes in the system state, such as UI state or database updates. This marks a shift in the evaluation paradigm from ``trace verification'' to ``result verification''\cite{spa_bench, he2025vitabenchbenchmarkingllmagents,yang2025macosworldmultilingualinteractivebenchmark}. 
This lays the foundation for a more realistic and accurate measurement of an agent's decision-making and execution abilities.

Despite these efforts, the development of Mobile GUI Agent benchmarks still faces several key challenges:
(1) Some benchmarks evaluate agents mainly on system applications or open-source apps~\cite{wang2024mobileagentbenchefficientuserfriendlybenchmark, android_world}, which differ substantially from everyday third-party mobile services.
(2) Benchmarks involving commercial apps often require running live apps on physical devices~\cite{spa_bench, zhang2024llamatouchfaithfulscalabletestbed}.
These environments are difficult to reproduce because recommendation algorithms, advertisements, account states, and dynamic page loading can change across runs.

As shown in Fig.\ref{fig_Device_Compare} (a), even for the same task, different agent attempts may face entirely different environmental states. This difference directly violates the fundamental reproducibility principle of benchmark design, affecting the comparability and reliability of the results.

In summary, building an evaluation environment that is both realistic and interactive, yet also controllable and reproducible\cite{better_bench, kapoor2023reformsreportingstandardsmachine}, has become a core need for Mobile GUI Agent research. Although existing work\cite{zhang2024mobileenvbuildingqualifiedevaluation} has noted the reproducibility issue, a general and efficient solution is still lacking. To this end, we propose AppSim-Bench: a new benchmark framework to systematically evaluate smartphone agents on everyday tasks.

AppSim-Bench addresses this need by constructing simulated apps with task-relevant UI flows and controllable backend data.
Rather than interacting with live commercial services, agents operate in sandboxed environments whose initial states and content can be reset before each run.
This design preserves realistic interaction patterns while removing major sources of evaluation noise from recommendation systems and dynamic content.

The benchmark is constructed through a coding-agent-assisted, human-verified workflow.
Researchers specify the task requirements, data schema, page hierarchy, and representative screenshots, while a coding agent generates candidate implementations under these constraints.
Human developers then inspect, execute, and iteratively correct the generated apps before they are used for evaluation.
This workflow reduces repeated implementation effort, but the benchmark contribution lies in the resulting controlled evaluation environment and task suite.

Our main contributions are as follows:
\begin{itemize}
    \item We introduce AppSim-Bench, a reproducible Mobile GUI Agent benchmark with 557 tasks across 17 simulated apps from Chinese and English mobile ecosystems.
    \item We design a deterministic evaluation protocol that separates benchmark specification, reference execution, and official scoring, covering state, answer, and hybrid task types.
    \item We evaluate 19 GUI agents and find that reliable mobile GUI execution remains far from solved: the strongest model completes only about half of the tasks, nearly one third of tasks are unsolved by any evaluated agent, and performance drops with task length, numerical constraints, and inefficient execution.
\end{itemize}
\newcommand{\cmark}{\checkmark}  %
\newcommand{\xmark}{$\times$}    %

\begin{table*}[!htbp]
  \centering
  \caption{
    Comparison of benchmark characteristics.
    AppSim-Bench bridges the gap between fidelity and reproducibility, pioneering the reproducible evaluation of high-frequency tasks derived from real-world mobile usage.
    In the context, ``Automated?'' indicates whether the evaluation is fully automated without human intervention,
    whereas ``Accurate?'' indicates whether the evaluation is robust and noise-free.
  }
    
  \label{tab:benchmark_compare}
  \resizebox{\textwidth}{!}{%
  \begin{tabular}{lcccccccc} %
    \toprule
    \textbf{Benchmark} 
    & \textbf{3rd-party App} 
    & \textbf{Reproducible} 
    & \textbf{\#Tasks} 
    & \textbf{\#Apps} 
    & \textbf{Avg. Tasks} 
    & \textbf{Automated?} 
    & \textbf{Accurate?} \\
    \midrule
    AndroidArena\cite{xing2024understandingweaknesslargelanguage}                     & \xmark & \xmark & 221 & 13 & 17.0 & \xmark & \xmark \\
    AndroidWorld\cite{android_world} & \cmark & \cmark & 116 & 20 & 5.8  & \xmark & \cmark \\
    LlamaTouch\cite{zhang2024llamatouchfaithfulscalabletestbed}                       & \cmark & \xmark & 496 & 57 & 8.7  & \xmark & \xmark \\
    B-MoCA\cite{lee2025benchmarkingmobiledevicecontrol}                           & \xmark & \xmark & 60  & 18 & 3.3  & \xmark & \xmark \\
    MobileAgentBench\cite{wang2024mobileagentbenchefficientuserfriendlybenchmark}                 & \xmark & \xmark & 100 & 10 & 10.0 & \xmark & \xmark \\
    SPA-BENCH\cite{spa_bench}        & \cmark & \xmark & 340 & 68 & 5.0  & \cmark & \xmark \\
    \midrule
    \textbf{AppSim-Bench (ours)}     & \cmark & \cmark & 557 &  17 & 32.8 & \cmark & \cmark \\
    \bottomrule
  \end{tabular}%
  }
\end{table*}

\section{Related Work} \label{sec:related_work}
\subsection{Mobile GUI Agent}

Benefiting from the rapid advances of vision-language models (VLMs) in visual understanding and reasoning, research on Mobile GUI Agents has progressed rapidly in recent years \cite{SeedVL, uitars_1, wang2025opencuaopenfoundationscomputeruse}. Leveraging these capabilities, GUI Agents can comprehend natural-language instructions, recognize the semantics and spatial layout of on-screen elements, and plan and execute multi-step actions accordingly.

Recent research has approached this direction from several perspectives:
(1) employing prompt engineering to adapt general-purpose large models for GUI interaction tasks directly \cite{ASSISTGUI, SeeAct};
(2) performing task-specific fine-tuning on subtasks such as interface understanding and element grounding \cite{AutoWebGLM, learnact, mobilevlm, agent_cpm_CAGUI, webarena};
and (3) constructing agent frameworks where specialized modules cooperate to form complete GUI workflows \cite{AgentS2, mobileagent_v3, AppAgent}.

Despite these advances, deploying current GUI Agents in real-world or production environments remains challenging due to issues such as semantic ambiguity of user intents and the complex functionality of mobile applications \cite{visualwebarena, liu2025llmpoweredguiagentsphone,nguyen2025guiagentssurvey}. These challenges highlight the urgent need for a benchmark that can reflect realistic usage scenarios while ensuring reproducibility and fairness.

\subsection{GUI Agent Benchmarks}
To systematically evaluate GUI Agents, early benchmarks often relied on static screenshots, recorded demonstrations, or human-annotated action traces \cite{agent_cpm_CAGUI, rawles2023androidwildlargescaledataset}. These settings are scalable and easy to standardize, but they often judge task completion by comparing the agent's action sequence with a human-annotated ``golden'' trace. Such trace-based evaluation struggles with the common multi-path problem in real Apps, where the same goal can be completed through different valid trajectories.

Subsequent studies began exploring dynamic evaluation frameworks, which execute interactive applications on virtual machines or physical devices to provide agents with testing environments that more closely resemble real-world usage scenarios \cite{bonatti2024windowsagentarenaevaluating, xie2024osworldbenchmarkingmultimodalagents, mialon2023gaia, deng2023mind2web}. For example, AndroidWorld\cite{android_world} offers agents a full execution interface and determines task completion by monitoring changes in system states. However, such approaches typically rely on system apps or open-source applications \cite{android_world, zhang2024mobileenvbuildingqualifiedevaluation}, as accessing backend states of commercial apps is often infeasible.

To evaluate on more everyday tasks, recent works have adopted physical-device-based evaluation, where agents interact directly with real mobile apps, and task success is assessed manually or via VLM-based evaluators \cite{he2025vitabenchbenchmarkingllmagents, spa_bench, learnact}.
While these attempts significantly increase the realism of evaluation, they still suffer from poor reproducibility and a lack of environmental control.
Recommendation algorithms, advertisement placements, and dynamic content updates can cause the same task to present different UI states at different times, thereby undermining the fairness and reliability of model comparisons. Table~\ref{tab:benchmark_compare} summarizes these benchmark differences in terms of app coverage, reproducibility, scale, and evaluation protocol. In summary, existing benchmarks face a clear trade-off between realism and reproducibility. This has become a major bottleneck for further research in GUI Agents.

A complementary line of work pursues the same goals of controllability and reproducibility through simulated app environments. AppWorld~\cite{trivedi-etal-2024-appworld} pioneered this direction with a controllable execution environment of 9 apps and state-based unit tests, and shares with us the principles of interactivity, reproducibility, and outcome-based verification. It targets a different agent paradigm, however: agents act by generating code against 457 APIs, whereas AppSim-Bench targets interface-based GUI agents that must ground every action in on-screen pixels through primitives such as tap, swipe, and type. The two settings therefore probe distinct capabilities, namely program synthesis and API composition as opposed to visual grounding and UI navigation. More recent platforms share our simulation-based premise but optimize for scalable training rather than evaluation, and accordingly adopt browser-hosted or synthesized implementations~\cite{wu2026mobilegymverifiablehighlyparallel, liu2026scalewobguidingguiagents}. AppSim-Bench instead runs on a native Android runtime and covers both Chinese and English ecosystems.

\begin{figure*}[!htbp]
   \begin{overpic}[width=\textwidth]{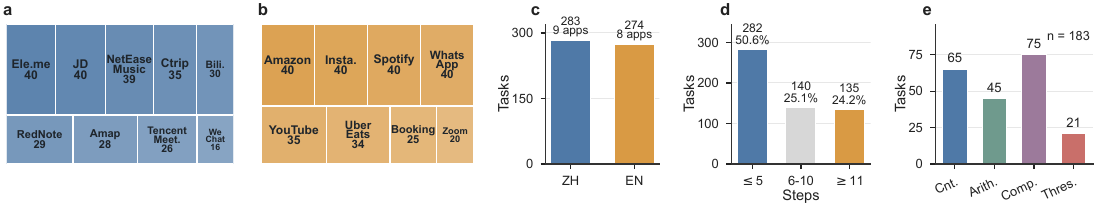} \small
   \end{overpic}
    \caption{
        Statistics of AppSim-Bench. 
        (a--b) App-level task coverage within the Chinese and English app ecosystems, where each rectangle denotes a simulated app and its task count.
        (c) Aggregate task coverage across both ecosystems.
        (d) Task distribution by human-annotated reference steps.
        (e) Distribution of numerical reasoning subtypes (subtypes are not mutually exclusive), here, ``Cnt.'', ``Arith.'', ``Comp.'', and ``Thres.''
        denote counting, arithmetic calculation, numeric comparison, and threshold filtering tasks.
    }
    \label{dataset}
\end{figure*}

\section{AppSim-Bench} \label{Method}
AppSim-Bench serves as a robust framework for evaluating mobile GUI agents within realistic, controllable, and reproducible environments. The benchmark comprises three core components:
(1) \textbf{Tasks}, which describe user intentions in natural language;
(2) \textbf{Simulated Apps}, a suite of task-relevant mobile application environments equipped with controllable backend data; and
(3) \textbf{Automated Evaluation} that determines task completion via deterministic outcome verification across state-based, answer-based, and hybrid task types.
The overall benchmark configuration is illustrated in Figure~\ref{fig_Device_Compare}(b), while a comprehensive summary of the task composition is provided in Figure~\ref{dataset}.

\paragraph{Task Construction} \label{tasks}
AppSim-Bench comprises 557 tasks across 17 simulated mobile applications from both Chinese and English application ecosystems, including WeChat, JD, RedNote, Tencent Meeting, Amap, Ctrip, Bilibili, Ele.me, NetEase Cloud Music, Amazon, Booking.com, Instagram, Spotify, Uber Eats, WhatsApp, YouTube, and Zoom.
These applications encompass a broad spectrum of high-frequency, ubiquitous mobile scenarios, spanning social communication, e-commerce, travel, food delivery, navigation, video sharing, music streaming, and video conferencing. 

To ensure realism and relevance, we conducted a user study with 15 volunteers to collect representative, real-world interactions, such as ``\textit{sending message to friend}'' and ``searching and purchasing a product online''. These raw interaction samples formed the basis of our initial task templates, which were subsequently refined and standardized through human annotation to yield natural, unambiguous, and machine-interpretable instructions.

Furthermore, we curate a specialized subset of \textbf{numerical-reasoning tasks}. These tasks demand that the agent extract numerical information from the graphical interface and execute mathematical or logical operations during interaction. We categorize these tasks into four non-mutually exclusive subtypes: counting, arithmetic calculation, numeric comparison, and threshold filtering. This design empowers AppSim-Bench to benchmark how effectively GUI agents perform multi-step interface navigation with quantitative reasoning.

\paragraph{Simulated App Construction}\label{pipeline}
We construct the simulated apps through a human-led, coding-agent-assisted process. Human developers specify the task requirements, backend data schemas, page hierarchies, and task-path screenshots for each app. Given these specifications, coding agents generate candidate implementations, including page layouts, widget bindings, navigation logic, and frontend behaviors. Human developers then inspect, execute, and correct the generated apps until the task-relevant UI flows and backend states match the benchmark specification.

The resulting apps reproduce task-relevant UI flows from real-world applications while using controllable backend data. 
This design guarantees that every task initiates from a deterministic state, and remains unaffected by environmental stochasticity induced by personalization, advertisements, recommendation updates, and dynamic online content. Complete construction details are provided in Appendix~\ref{app:simulated-app-construction}, and the quality-control procedure in Appendix~\ref{app:construction-qc}.

\paragraph{Automated Evaluation}\label{Evaluation}
AppSim-Bench uses deterministic outcome evaluation rather than predefined action-path matching. Following AndroidWorld\cite{android_world}, the evaluator checks whether the final outcome satisfies the task goal after an agent completes an interaction. We use three verification types:
\textbf{(1) State tasks} require a visible or backend state change, such as sending a message or placing an order.
\textbf{(2) Answer tasks} require the agent to retrieve information and return a structured answer, such as a count, price, or selected option.
\textbf{(3) Hybrid tasks} require both a state change and a structured answer.
The evaluator applies the corresponding deterministic check for each task type.
Detailed schemas and examples are provided in Appendix~\ref{app:evaluation-details}, and a representative agent trajectory on an answer task is shown in Appendix~\ref{app:case-study}.

\paragraph{Statistics of AppSim-Bench}\label{Statistics}

As summarized in Figure~\ref{dataset}, AppSim-Bench comprises 557 tasks across 17 simulated apps from Chinese and English app ecosystems. These two ecosystems are approximately balanced, with 283 tasks from 9 Chinese apps and 274 tasks from 8 English apps. We categorize tasks by the human-annotated reference steps required for completion: 282 tasks require at most 5 steps, 140 require 6--10 steps, and 135 require at least 11 steps. Among all tasks, 183 are annotated as numerical-reasoning tasks, with subtype labels for counting, arithmetic calculation, numeric comparison, and threshold filtering.
Full statistics are reported in Appendix~\ref{sec:app_sim}.

\begin{table*}[t]
\centering
\caption{Accuracy summary on AppSim-Bench across overall performance, task length, and numerical-reasoning subtype. ``ZH Apps'' and ``EN Apps'' denote task-weighted accuracy over the Chinese and English app groups. Task-length groups are based on the human-annotated task step count $\#S$. ``Cnt.'', ``Arith.'', ``Comp.'', and ``Thres.'' denote counting, arithmetic calculation, numeric comparison, and threshold filtering tasks. \textbf{Bold} and \underline{underlining} indicate the highest and second-highest accuracies in each column. {\textcolor{red!80!black}{$\downarrow$}} and {\textcolor{green!60!black}{$\uparrow$}} indicate accuracy variations relative to the adjacent shorter task-length bucket. {\textcolor{red!80!black}{\textsuperscript{$\dagger$}}} highlights the most substantial performance degradation within each adjacent-bucket comparison. Note that numerical reasoning subtypes are not mutually exclusive.}
\label{tab:SR_overall_length_subtask_merged}
\label{tab:overall-acc-summary}
\resizebox{\textwidth}{!}{%
\begin{tabular}{lcccccccccc}
\toprule
\multirow{2}{*}{Agent Model} &
\multicolumn{3}{c}{Overall Benchmark} &
\multicolumn{3}{c}{Task Length} &
\multicolumn{4}{c}{Numerical Reasoning Subtype} \\
\cmidrule(lr){2-4}\cmidrule(lr){5-7}\cmidrule(lr){8-11}
& Overall & ZH Apps & EN Apps
& $\#S_{\leq 5}$ & $\#S_{6\text{-}10}$ & $\#S_{\geq 11}$
& Cnt. & Arith. & Comp. & Thres. \\
\midrule
\multicolumn{11}{c}{\textit{General-purpose Agents}} \\
\midrule
Claude-Opus-4.7 & \textbf{50.27\%} & \textbf{60.07\%} & \underline{40.15\%} & \underline{58.87\%} & \textbf{54.29\%} \down{4.58} & 28.15\% \down{26.14}\textsuperscript{\textcolor{red!80!black}{$\dagger$}} & \textbf{52.31\%} & \underline{55.56\%} & \textbf{52.00\%} & \textbf{42.86\%} \\
Gemini-3-Pro & \underline{49.91\%} & \textbf{60.07\%} & 39.42\% & 56.38\% & \underline{52.86\%} \down{3.52} & \textbf{33.33\%} \down{19.53} & \underline{46.15\%} & 42.22\% & 49.33\% & 23.81\% \\
GPT-5.5 & 49.73\% & 58.66\% & \textbf{40.51\%} & \textbf{59.22\%} & 52.14\% \down{7.08} & 27.41\% \down{24.73} & 44.62\% & \textbf{57.78\%} & \textbf{52.00\%} & \underline{33.33\%} \\
Gemini-3.1-Pro & 49.55\% & \underline{59.36\%} & 39.42\% & 56.74\% & \underline{52.86\%} \down{3.88} & \underline{31.11\%} \down{21.75} & 43.08\% & 42.22\% & \underline{50.67\%} & 23.81\% \\
Doubao-Seed-1.6 & 39.14\% & 50.88\% & 27.01\% & 45.74\% & 43.57\% \down{2.17} & 20.74\% \down{22.83} & 23.08\% & 22.22\% & 36.00\% & 9.52\% \\
GPT-5 & 33.39\% & 37.81\% & 28.83\% & 43.62\% & 31.43\% \down{12.19} & 14.07\% \down{17.36} & 27.69\% & 35.56\% & 40.00\% & 4.76\% \\
Doubao-Seed-1.8 & 32.14\% & 34.63\% & 29.56\% & 43.26\% & 27.86\% \down{15.40} & 13.33\% \down{14.53} & 23.08\% & 31.11\% & 28.00\% & 4.76\% \\
Qwen-3.6-Plus & 31.42\% & 39.58\% & 22.99\% & 39.01\% & 31.43\% \down{7.58} & 15.56\% \down{15.87} & 23.08\% & 13.33\% & 28.00\% & 9.52\% \\
Doubao-Seed-2.0 & 29.26\% & 36.40\% & 21.90\% & 37.94\% & 28.57\% \down{9.37} & 11.85\% \down{16.72} & 27.69\% & 28.89\% & 20.00\% & 14.29\% \\
GPT-5.4 & 29.08\% & 34.63\% & 23.36\% & 38.30\% & 25.71\% \down{12.59} & 13.33\% \down{12.38} & 24.62\% & 24.44\% & 30.67\% & 9.52\% \\
Qwen-3.6-Flash & 28.19\% & 38.87\% & 17.15\% & 38.30\% & 25.71\% \down{12.59} & 9.63\% \down{16.08} & 20.00\% & 24.44\% & 26.67\% & 4.76\% \\
Claude-Sonnet-4.6 & 27.29\% & 31.10\% & 23.36\% & 36.17\% & 27.86\% \down{8.31} & 8.15\% \down{19.71} & 29.23\% & 33.33\% & 26.67\% & 14.29\% \\
Claude-Haiku-4.5 & 22.80\% & 22.61\% & 22.99\% & 34.04\% & 15.71\% \down{18.33}\textsuperscript{\textcolor{red!80!black}{$\dagger$}} & 6.67\% \down{9.04} & 20.00\% & 22.22\% & 20.00\% & 9.52\% \\
Claude-Sonnet-4.5 & 21.01\% & 24.03\% & 17.88\% & 31.91\% & 14.29\% \down{17.62} & 5.19\% \down{9.10} & 23.08\% & 26.67\% & 21.33\% & 14.29\% \\
Gemini-2.5-Pro & 17.24\% & 21.20\% & 13.14\% & 26.24\% & 14.29\% \down{11.95} & 1.48\% \down{12.81} & 23.08\% & 17.78\% & 17.33\% & 9.52\% \\
\midrule
\multicolumn{11}{c}{\textit{GUI-specialized Agents}} \\
\midrule
MobileAgent-v3.5 & 31.42\% & 35.34\% & 27.37\% & 41.49\% & 26.43\% \down{15.06} & 15.56\% \down{10.87} & 18.46\% & 24.44\% & 32.00\% & 4.76\% \\
UI-TARS-1.5-7B & 22.62\% & 31.10\% & 13.87\% & 30.14\% & 22.86\% \down{7.28} & 6.67\% \down{16.19} & 20.00\% & 8.89\% & 16.00\% & 4.76\% \\
AgentCPM-GUI & 10.05\% & 9.19\% & 10.95\% & 17.02\% & 5.71\% \down{11.31} & 0.00\% \down{5.71} & 3.08\% & 2.22\% & 5.33\% & 4.76\% \\
V-Droid & 9.87\% & 13.43\% & 6.20\% & 15.25\% & 7.86\% \down{7.39} & 0.74\% \down{7.12} & 15.38\% & 11.11\% & 9.33\% & 0.00\% \\
\bottomrule
\end{tabular}%
}
\end{table*}

\section{Experiments} \label{sec:Experiment}
\subsection{Experimental Setup} \label{sec:exp_setup}

\paragraph{Baselines.}
We evaluate 15 powerful general-purpose agents from five model families on AppSim-Bench: GPT-(5.5/5.4/5), Claude-(Opus-4.7/Sonnet-4.6/Sonnet-4.5/Haiku-4.5), Gemini-(3.1/3/2.5)-Pro, Qwen-3.6-(Plus/Flash), and Doubao-Seed-(2.0/1.8/1.6).
Qwen and Doubao are evaluated with their official GUI agent prompts and tool schemas. For Claude, GPT, and Gemini, we adapt M3A~\cite{android_world}, retaining its mobile observation and action interface, but use a single-model version to match the setting of Qwen and Doubao.
We further evaluate 4 GUI-specialized agents that are post-trained for GUI interaction: MobileAgent-v3.5~\cite{mobileagent_v3}, UI-TARS-1.5-7B~\cite{uitars_1}, AgentCPM-GUI~\cite{agent_cpm_CAGUI}, and V-Droid~\cite{dai2026advancingmobileguiagents}, each integrated with its own official GUI prompt and action schema. Among them, MobileAgent-v3.5 is the only multi-agent system. This yields 19 evaluated agents in total.

\paragraph{Evaluation Protocol.}
All experiments are conducted on devices running Android 13 (API 33) at a screen resolution of $1080 \times 1920$. Before each task begins, we clear the app state, restore the task-specific initial state, and reset the agent history. At each step, the agent receives the task instruction, current screen observation, and a concise action history summarizing previous operations within the same task. We set the temperature to $0$ for all evaluated agents, and cap each run at 50 steps. More details are reported in Appendix~\ref{app:agent-prompts}.

\paragraph{Metrics.}
We evaluate all baseline agents' performance using a single binary metric: the agent receives a score of $1$ when the final goal condition is satisfied, and $0$ otherwise.
For state tasks, this score is produced by checking whether the final app state satisfies the predefined goal condition.
For the answer and hybrid evaluation cases reported in this paper, we manually audited the structured-answer extraction outputs used for scoring and found no extraction errors in the audited cases.
For each model evaluated on $\mathcal{T}$, let $y_i \in \{0,1\}$ denote the score on task $i$, $a_i$ denote the number of valid actions, and $h_i$ denote the human-annotated reference steps.
We define accuracy (ACC) and average actions (AvgAct) as,
\[
    \begin{aligned}
        \mathrm{ACC} = \frac{1}{|\mathcal{T}|}\sum_{i\in\mathcal{T}} y_i,
        \mathrm{AvgAct} = \frac{1}{|\mathcal{T}|}\sum_{i\in\mathcal{T}} a_i.
    \end{aligned}
\]
To measure efficiency of successfully completed tasks, we define Action Overhead (ActOH) as, 
\[
    \mathrm{ActOH}=\frac{1}{\sum_{i\in\mathcal{T}}y_i}    \sum_{i\in\mathcal{T}} y_i\frac{a_i}{h_i},
\]
where lower values indicate trajectories closer to the human reference length.

\begin{figure}[!h]
   \centering
   \includegraphics[width=\columnwidth]{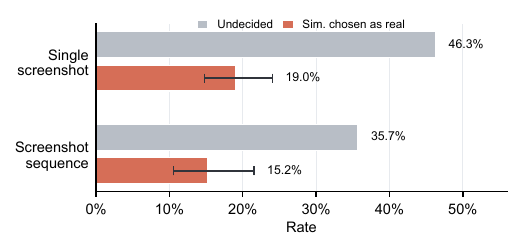}
   \caption{Blind human evaluation of visual realism. Participants compared real-app content with the corresponding simulated-app content under two settings: single screenshots and screenshot sequences from continuous interactions. ``Undecided'' is computed over all judgments. ``Sim. chosen as real'' is computed over judgments with an explicit A/B choice; error bars denote Wilson 95\% confidence intervals.}
   \label{fig:ab-realism-study}
\end{figure}

\subsection{Human Validation of Visual Realism} \label{sec:exp_fidelity}
To verify whether AppSim-Bench preserves sufficient visual realism for agent evaluation, seven annotators conducted a blind A/B study over 11 sessions, comparing matched real and simulated assets across both static screenshots and dynamic interaction sequences. For each trial, evaluators identified the authentic interface or selected "undecided."

The study demonstrates the high visual fidelity of our simulated applications. For static screenshots (510 judgments), 46.3\% were deemed "undecided," and the simulation was mistaken for the real app in 19.0\% of the 274 decided cases. For dynamic sequences (255 judgments), the "undecided" rate was 35.7\%, with the simulation selected as real in 15.2\% of the 164 decided cases. These results provide empirical validation that our simulated environments are visually non-trivial to distinguish from real-world services.

Figure~\ref{fig_PageTree} in the Appendix compares the page-transition topologies of a real app and its simulator.
The simulator faithfully preserves task-relevant pages and transitions, while strategically pruning extraneous real-app states that fall outside the scope of our benchmark tasks.
Appendix~\ref{app:placeholder-residual} quantifies the residual effect of this pruning: under a pessimistic upper bound it accounts for at most 4.28\% of measured success.

\begin{figure}[!h]
    \centering
    \includegraphics[width=1.0\columnwidth]{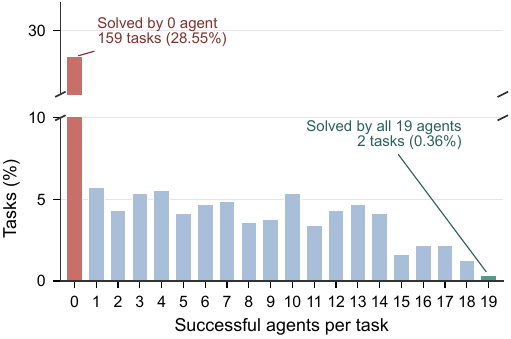}
    \caption{Cross-agent task solvability. The chart shows the proportion of tasks solved by $k$ agents.}
    \label{fig:solved-by-k-distribution}
\end{figure}

\begin{table*}[t]
    \centering
    \caption{Action efficiency and budget-exhaustion behavior. Arrows indicate the preferred performance direction. AvgAct is the average number of valid actions across all tasks. ActOH (Action Overhead) is computed on successful tasks as the ratio of valid actions to human-annotated reference steps. Budg. Exh. counts trajectories that exhaust the predefined 50-step interaction budget. Budg. ACC reports accuracy among budget-exhausted trajectories. Rep.$\geq$3 and Rep.$\geq$5 specify the count and percentage of budget-exhausted trajectories that contain at least three or five consecutive repetitions of the same normalized action signature.}
    \label{tab:action-efficiency-budget}
    \resizebox{\textwidth}{!}{%
    \begin{tabular}{lrrrrrrr}
        \toprule
        Agent Model & ACC $\uparrow$ & AvgAct $\downarrow$ & ActOH $\downarrow$ & Budg. Exh. $\downarrow$ & Budg. ACC $\uparrow$ & Rep.$\geq$3 $\downarrow$ & Rep.$\geq$5 $\downarrow$ \\
        \midrule
        \multicolumn{8}{c}{\textit{General-purpose Agents}} \\
        \midrule
        Claude-Opus-4.7   & 50.27\% & 14.01        & 1.98       & 33          & 0.00\%          & 27 (81.82\%)             & 20 (60.61\%)             \\
        Gemini-3-Pro      & 49.91\% & 10.79        & 1.69       & 14          & 0.00\%          & 5 (35.71\%)              & 3 (21.43\%)              \\
        GPT-5.5           & 49.73\% & 15.04        & 2.09       & 45          & 0.00\%          & 27 (60.00\%)             & 15 (33.33\%)             \\
        Gemini-3.1-Pro    & 49.55\% & 10.69        & 1.80       & 19          & 10.53\%         & 7 (36.84\%)              & 3 (15.79\%)              \\
        Doubao-Seed-1.6   & 39.14\% & 15.59        & 1.73       & 85          & 1.18\%          & 74 (87.06\%)             & 52 (61.18\%)             \\
        GPT-5             & 33.39\% & 23.15        & 2.94       & 136         & 0.00\%          & 105 (77.21\%)            & 61 (44.85\%)             \\
        Doubao-Seed-1.8   & 32.14\% & 9.71         & 1.60       & 44          & 0.00\%          & 36 (81.82\%)             & 27 (61.36\%)             \\
        Qwen-3.6-Plus     & 31.42\% & 12.66        & 1.62       & 23          & 0.00\%          & 11 (47.83\%)             & 9 (39.13\%)              \\
        Doubao-Seed-2.0   & 29.26\% & 10.76        & 1.81       & 46          & 0.00\%          & 33 (71.74\%)             & 28 (60.87\%)             \\
        GPT-5.4           & 29.08\% & 10.88        & 1.56       & 16          & 0.00\%          & 12 (75.00\%)             & 7 (43.75\%)              \\
        Qwen-3.6-Flash    & 28.19\% & 13.30        & 2.10       & 38          & 0.00\%          & 27 (71.05\%)             & 23 (60.53\%)             \\
        Claude-Sonnet-4.6 & 27.29\% & 29.80        & 3.11       & 253         & 0.40\%          & 222 (87.75\%)            & 171 (67.59\%)            \\
        Claude-Haiku-4.5  & 22.80\% & 31.63        & 4.07       & 221         & 0.00\%          & 208 (94.12\%)            & 162 (73.30\%)            \\
        Claude-Sonnet-4.5 & 21.01\% & 23.31        & 3.30       & 93          & 3.23\%          & 81 (87.10\%)             & 55 (59.14\%)             \\
        Gemini-2.5-Pro    & 17.24\% & 22.10        & 2.38       & 106         & 0.00\%          & 74 (69.81\%)             & 40 (37.74\%)             \\
        \midrule
        \multicolumn{8}{c}{\textit{GUI-specialized Agents}} \\
        \midrule
        MobileAgent-v3.5  & 31.42\% & 16.79        & 2.02       & 83          & 1.20\%          & 69 (83.13\%)             & 40 (48.19\%)             \\
        UI-TARS-1.5-7B    & 22.62\% & 24.76        & 1.67       & 226         & 0.00\%          & 156 (69.03\%)            & 138 (61.06\%)            \\
        AgentCPM-GUI      & 10.05\% & 24.82        & 1.71       & 242         & 0.00\%          & 203 (83.88\%)            & 198 (81.82\%)            \\
        V-Droid           & 9.87\%  & 33.30        & 4.60       & 313         & 0.32\%          & 313 (100.00\%)           & 295 (94.25\%)            \\
        \midrule
        Summary               & 30.76\% & 18.58        & 2.15       & 2036        & 0.44\%          & 1690 (83.01\%)           & 1347 (66.16\%)           \\
        \bottomrule
    \end{tabular}
    }
\end{table*}

\subsection{Overall Performance} \label{sec:exp_overall}

Table~\ref{tab:SR_overall_length_subtask_merged} consolidates three complementary perspective of model performance: overall accuracy, task-length robustness, and subtype accuracy in numerical-reasoning. 
The overall results underscore that reliable autonomous GUI task execution remains an elusive goal for state-of-the-art agents. The frontier model, Claude-Opus-4.7, reaches a ceiling of 50.27\% overall accuracy. 
Gemini-3-Pro, GPT-5.5, and Gemini-3.1-Pro follow closely, yet all remain below 50\%. 
The remaining models form a substantially lower tier. For example, GPT-5 achieves 33.39\% overall accuracy, whereas Gemini-2.5-Pro reaches only 17.24\%. This stratification demonstrates that AppSim-Bench captures both granular deltas among frontier models and wider capability gaps among weaker baselines.
GUI-specialized post-training does not close this gap. The strongest such agent, MobileAgent-v3.5, reaches 31.42\%, and the other three remain below 23\%. The ceiling on AppSim-Bench is therefore set by the task demands rather than by the absence of GUI-specific post-training.
The group-level evaluations further demonstrate that agent performance is non-uniform across application ecosystems. Leading models reach approximately 59--60\% accuracy on the Chinese applications, yet only about 39--41\% on the English applications. 
Superior performance within one linguistic ecosystem does not guarantee comparable reliability in the other. The app-level results in Tables~\ref{tab:app-acc-zh} and~\ref{tab:app-acc-en} further demonstrate substantial variation across different applications.

\begin{figure*}[t]
    \centering
    \includegraphics[width=\textwidth]{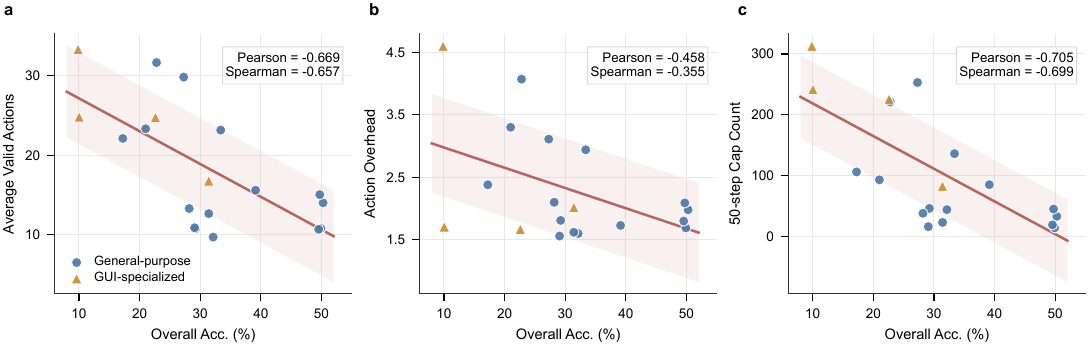}
    \caption{Accuracy-efficiency relationships across agents. Each point denotes one agent. Panels show overall accuracy vs. (a) average valid actions, (b) Action Overhead, and (c) budget-exhausted trajectories. Pearson and Spearman correlations are shown in each panel.}
    \label{fig:efficiency-analysis}
\end{figure*}

Figure~\ref{fig:solved-by-k-distribution} further examines task-level coverage across models. The distribution shows that 28.55\% of tasks are not solved by any model, indicating a substantial set of shared failure cases. At the other end, only 0.36\% of tasks are solved by all models included in this analysis. This pattern suggests that current GUI agents do not merely differ in overall accuracy. They also share systematic blind spots on a non-trivial portion of AppSim-Bench.
Appendix~\ref{app:unsolved-failure-modes} categorizes these shared failures into three capability gaps.

\subsection{Task Length as a Long-horizon Bottleneck} \label{sec:exp_task_length}

The middle block of Table~\ref{tab:SR_overall_length_subtask_merged} groups tasks by the human-annotated number of reference steps required for completion.
It reports model accuracy on three task-length buckets, $\#S_{\leq 5}$, $\#S_{6\text{-}10}$, and $\#S_{\geq 11}$.
A clear pattern emerges: accuracy consistently decreases as the required number of reference steps increases.
Every evaluated model performs worse on $\#S_{6\text{-}10}$ tasks than on $\#S_{\leq 5}$ tasks, and worse again on $\#S_{\geq 11}$ tasks.
For example, GPT-5 drops from 43.62\% to 31.43\% and then to 14.07\%.
The top-performing model also varies across task lengths: GPT-5.5 leads on $\#S_{\leq 5}$ tasks, Claude-Opus-4.7 leads on $\#S_{6\text{-}10}$ tasks, and Gemini-3-Pro leads on $\#S_{\geq 11}$ tasks.
However, the strongest long-task result is only 33.33\%, indicating that no evaluated model reliably handles the longest workflows.

The magnitude of the drop further shows where degradation is most severe.
Claude-Haiku-4.5 has the largest decrease from $\#S_{\leq 5}$ to $\#S_{6\text{-}10}$, falling by 18.33\%.
Claude-Opus-4.7 has the largest decrease from $\#S_{6\text{-}10}$ to $\#S_{\geq 11}$, falling by 26.14\%.
Overall, longer interaction consistently reduces model reliability.
In longer workflows, agents must track completed operations, interpret the current screen against the original instruction, and prevent intermediate errors from accumulating.
The consistent degradation indicates that current agents still struggle to maintain coherent progress over extended mobile interactions.
The same monotonic decline holds for the GUI-specialized agents, and is even steeper: AgentCPM-GUI and V-Droid fall to 0.00\% and 0.74\% on $\#S_{\geq 11}$ tasks. The long-horizon bottleneck is thus independent of model class.

\subsection{Task Type Analysis} \label{sec:exp_task_type}

Sections~\ref{sec:exp_overall} and~\ref{sec:exp_task_length} show two sources of difficulty in AppSim-Bench: aggregate reliability gaps and long-horizon degradation.
We next examine whether performance also varies with the type of reasoning required by the task.
The right block of Table~\ref{tab:SR_overall_length_subtask_merged} focuses on numerical reasoning tasks and breaks them down by subtype.
The leading models in the overall benchmark remain strong, but their relative advantages become more subtype-specific.
Claude-Opus-4.7 leads on counting and threshold filtering, while GPT-5.5 leads on arithmetic calculation.
Gemini-3-Pro and Gemini-3.1-Pro remain competitive on numeric comparison, but they are weaker on threshold filtering.
Thus, task type exposes capability differences that are partly hidden by aggregate benchmark scores.

The subtype breakdown further shows that the difficulty is not limited to arithmetic alone.
On arithmetic calculation, GPT-5.5 reaches 57.78\%, and Claude-Opus-4.7 reaches 55.56\%.
Counting and numeric comparison are also relatively stable for the leading models, with best accuracies around 52\%.
Threshold filtering is the main exception.
Claude-Opus-4.7 reaches 42.86\%, GPT-5.5 reaches 33.33\%, and most other models remain below 15\%.
This indicates that applying a numeric cutoff during UI search or selection is substantially harder than computing or comparing an isolated value.

Lower-tier models show a sharper collapse on the same subtype, with several systems reaching only 4.76\% or 9.52\%.
Overall, numerical reasoning in AppSim-Bench is better understood as a composite capability.
It requires value extraction, arithmetic manipulation, comparison, and constraint maintenance under interactive GUI conditions.
A manual audit reported in Appendix~\ref{app:numerical-error-decomposition} attributes these failures almost entirely to the reasoning stage: only 1 of 1,287 audited execution steps contained a perception or extraction error.

\subsection{Efficiency Analysis} \label{sec:exp_efficiency}

We next examine whether final task success is accompanied by efficient execution.
Table~\ref{tab:action-efficiency-budget} reports three complementary indicators: average valid actions, Action Overhead, and budget exhaustion.
These metrics distinguish accurate agents that act efficiently from agents that rely on long or repetitive trajectories.

Figure~\ref{fig:efficiency-analysis} shows that accuracy is negatively associated with all three inefficiency indicators.
Models with higher accuracy tend to use fewer actions on average, with Pearson and Spearman correlations of -0.669 and -0.657.
The same pattern holds for Action Overhead, with correlations of -0.458 and -0.355.
Budget exhaustion is also negatively correlated with accuracy, with correlations of -0.705 and -0.699.
Thus, low accuracy is not merely a final-state failure.
It is also associated with longer action sequences, less human-like successful trajectories, and more frequent budget exhaustion.
The association is weakest for Action Overhead, because the GUI-specialized agents produce short trajectories on the tasks they do solve while solving far fewer of them.

At the model level, the strongest systems combine relatively high accuracy with limited budget exhaustion.
Claude-Opus-4.7 reaches 50.27\% accuracy with 33 budget-exhausted trajectories, and the lower-accuracy model Claude-Haiku-4.5 reaches 22.80\% accuracy with 221.
Budget-exhausted trajectories are almost never successful, with an accuracy of only 0.44\%.
This makes budget exhaustion a strong empirical marker of failed execution.

Repeated-action statistics provide a finer-grained view of budget-exhausted runs.
Across all models, 1,690 such trajectories contain at least three consecutive repetitions of the same normalized action signature, and 1,347 contain at least five.
This pattern is especially pronounced in lower-accuracy models.
These statistics should not be interpreted as exact loop counts, because repeated actions may be valid when the screen state changes.
Nevertheless, consecutive repetition within budget-exhausted trajectories indicates limited progress despite continued interaction.

\section{Conclusion}\label{sec:Conclusion}

We present AppSim-Bench, a reproducible benchmark designed to evaluate mobile GUI agents within realistic and controllable mobile environments. 
AppSim-Bench leverages 
controllable backend data and deterministic outcome verification to mitigate the environmental volatility inherent in live commercial applications.
Extensive evaluations across 19 prominent GUI agents underscore that reliable autonomous mobile execution remains an unresolved challenge. The leading model completes only 50.27\% of the tasks, while 28.55\% remain entirely unsolved by any agent. Performance drops sharply within longer workflows and numerical reasoning tasks. Elevated action overhead and frequent interaction budget exhaustion also present as a major inefficiency. The evaluation highlights a critical need for improved long-horizon state tracking, robust numerical constraint handling, and sophisticated recovery mechanisms from stalled interactions.

\section*{Limitations}
AppSim-Bench prioritizes reproducibility by evaluating GUI agents in controllable simulated environments rather than live commercial apps. This design removes major sources of evaluation noise, including personalization, advertisements, network variation, account-specific content, and continuously changing online services. These factors, however, remain important for understanding agent behavior in open-ended mobile ecosystems. Thus, AppSim-Bench provides a stable basis for comparative evaluation, while complementary real-world studies are still needed to assess deployment-time robustness. 
The current benchmark covers a broad but incomplete subset of mobile interaction scenarios. Although it includes 557 tasks across 17 Chinese and English apps, future work should further expand task coverage, especially toward cross-app workflows, real-time interaction, accessibility-oriented tasks, and security-sensitive scenarios.

Realism is approximated per dimension rather than wholesale. We reproduce the visual appearance, in-page interaction, and navigation logic of the retained pages, but deliberately simplify four others: functional breadth, since we cover the task-relevant subset rather than an app's full feature set; dynamic and real-time content, such as live feeds and personalized recommendations; system-level factors, such as network latency and loading failures; and data scale, since manually constructed backend data need not reflect the long-tail histories of real users. These are trade-offs made in exchange for deterministic verification, and they are largely orthogonal to whether an agent can bring about a target state change.

Our fidelity and quality evidence is likewise bounded. Behavioral fidelity is measured indirectly, through the placeholder residual of Appendix~\ref{app:placeholder-residual} rather than a direct coverage ratio of implemented interactive elements. Construction quality rests on deterministic verification and multi-round human inspection (Appendix~\ref{app:construction-qc}) rather than standardized inter-annotator agreement or a residual bug rate. The error decomposition of Appendix~\ref{app:numerical-error-decomposition} is a deep audit of a single app, so its near-zero extraction rate remains to be confirmed across ecosystems.

\section*{Ethics Statement}
AppSim-Bench is intended for non-commercial research on mobile GUI agent evaluation. Its simulated apps are built for controlled benchmarking and are not connected to commercial services, user accounts, payment systems, or private user data. All backend data are manually defined or synthetic, and no real user records are used. 
The benchmark does not imply endorsement by, affiliation with, or substitution for any commercial product. App names are used only to describe the simulated interaction scenarios. The goal is reproducible research evaluation, not reproduction of proprietary services, redistribution of commercial app functionality, or automation of real-world services. 
AppSim-Bench is extensible, allowing researchers to customize tasks, app logic, and backend data. Such extensions should include privacy and safety safeguards. Custom backend data should not contain personal information or other sensitive records. Future releases and derivative uses should remain restricted to sandboxed research and evaluation settings.

\section*{Acknowledgments}

This work was supported by the National Natural Science Foundation of China
(Grant No. 62302186) and the Fundamental Research Funds for the Central
Universities (Grant Nos. CCNU25ai017 and 2026CXGR089).

We sincerely thank Agentic Labs at Acrab AI for its support of this work.
We also thank Luyao Chang, Xinzhu Du, Sirui Tang, and De Wu for their time
and effort in developing the simulated apps.

\bibliography{main}

\appendix
\section{Simulated App Construction Details}\label{app:simulated-app-construction}
\begin{figure*}[!htbp]
   \begin{overpic}[width=\textwidth]{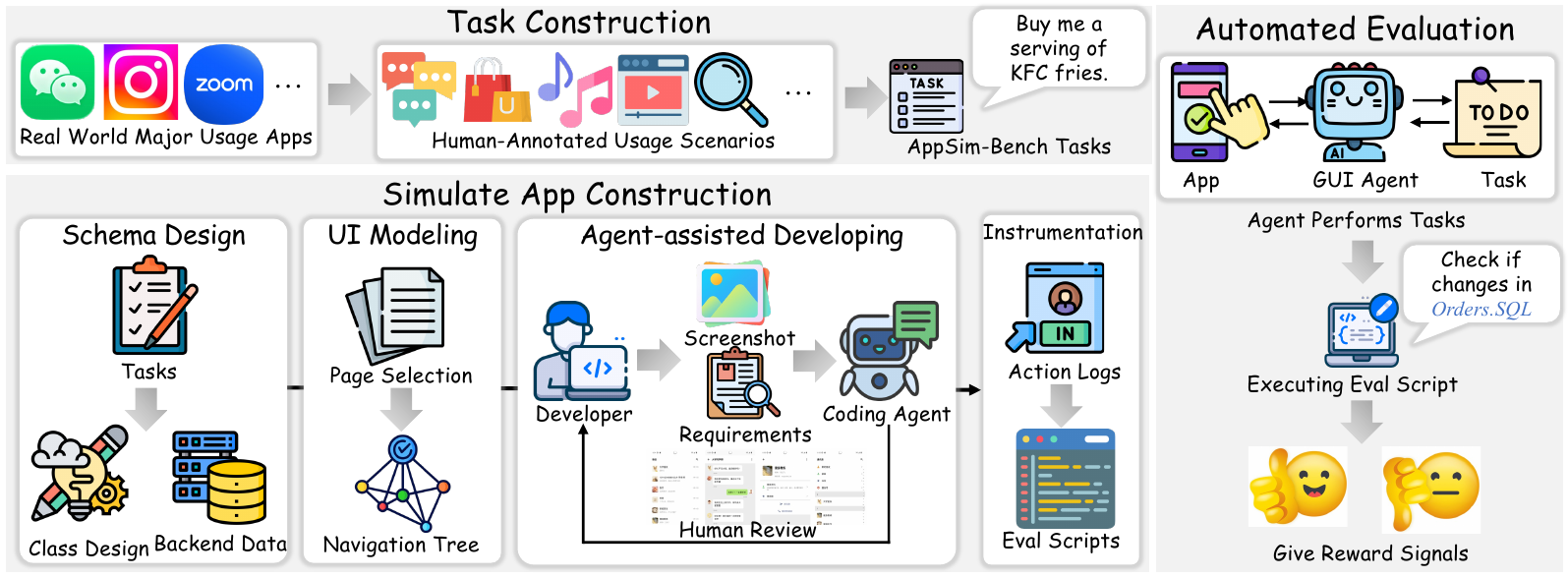} \small
   \end{overpic}
    \caption{
    Supplementary overview of simulated app construction in AppSim-Bench.
    The process specifies task requirements, backend data schemas, page structures, and verification signals for each simulated app.
    Coding agents are used as implementation assistance under human specification and inspection.
    }
    \label{app:sim-app-construction-overview}
\end{figure*}

Figure~\ref{app:sim-app-construction-overview} summarizes the construction process for the simulated apps used in AppSim-Bench.
The process has four main steps.
\begin{itemize}
    \item \textbf{Schema Design}: Define task-aware backend schemas that contain the entities and attributes required for execution and evaluation (Appendix ~\ref{app:simulated-app-construction_step_1}).
    \item \textbf{UI Modeling}: Specify task-relevant UI flows and page hierarchies for each app (Appendix ~\ref{app:simulated-app-construction_step_2}).
    \item \textbf{Agent-assisted Developing}: Human developers use coding agents to generate candidate implementations, then inspect and correct the apps  (Appendix ~\ref{app:simulated-app-construction_step_3}).
    \item \textbf{Instrumentation}: Add verification scripts, structured logs, and state records for deterministic evaluation (Appendix ~\ref{app:simulated-app-construction_step_4}).
\end{itemize}

\subsection{Schema Design}\label{app:simulated-app-construction_step_1}
For each simulated app, we define backend data schemas according to the benchmark tasks it must support.
We identify task-relevant entities and attributes, and omit fields that are not needed for task execution or evaluation.
For example, in modeling the ``User'' class for WeChat, we retain attributes such as \textit{name}, \textit{avatar}, and \textit{status}, while excluding fields such as \textit{last\_login\_time} and \textit{device\_info}.
This selective design keeps the app state sufficient for benchmark evaluation without adding unrelated backend complexity.

\subsection{UI Modeling}\label{app:simulated-app-construction_step_2}
For each app, we specify a task-relevant page hierarchy and interaction paths based on the app's functional structure.
For instance, when simulating WeChat, we organize the app into four main tabs, \textit{Chat}, \textit{Contact}, \textit{Discover}, and \textit{Me}, and define the subpages and navigation paths needed by the benchmark tasks.
The page hierarchy constrains the simulated UI flows and helps ensure that the implemented interactions support the intended tasks.

\subsection{Agent-assisted Developing}\label{app:simulated-app-construction_step_3}
Coding agents are used as implementation assistance, not as an autonomous benchmark-construction method.
In our implementation, developers provide task specifications, page hierarchies, and representative screenshots to guide candidate app implementations.
The generated layouts, widget bindings, navigation logic, and frontend behaviors are then inspected and corrected by human developers.
Before evaluation, each app is executed and checked to ensure that the task-relevant UI flows and backend states match the benchmark specification.

\subsection{Instrumentation}\label{app:simulated-app-construction_step_4}
Each simulated app records task-relevant state changes through structured logs or backend fields.
For example, WeChat maintains chat logs for message-sending tasks, while Ele.me records click logs and order states for ordering tasks.
These records allow the evaluator to verify task outcomes from a controlled initial state.
Because the backend data are predefined, the evaluation is not affected by recommendation algorithms, advertisements, account-specific content, or changing online services.

\section{Evaluation Schemas and Examples}\label{app:evaluation-details}

For state tasks, each task is represented as $(S_0, G, S_t)$, where $S_0$ is the initial state, $G$ is the goal condition, and $S_t$ is the final state after execution.
During execution, the app records structured logs or backend fields.
The evaluator compares $S_t$ with $G$ and marks success only when target fields show the expected changes, such as updated chat history, order status, or like count.

For answer tasks, each task defines an answer schema with required fields, data types, and format constraints.
If an agent returns free-form text, a shared reference executor extracts a structured answer under this schema.
The evaluator then performs schema validation, canonicalization, and deterministic matching against the gold answer.
This scoring stage contains no LLM calls.

For hybrid tasks, both the state-transition check and the answer-schema check must pass.
All reported baselines use the same reference executor.
Users may also bypass it by making agents directly output schema-conformant JSON.

To verify that the answer extraction step does not introduce evaluation noise into the reported results, we manually checked the structured-answer extraction outputs used to score the answer and hybrid evaluation cases reported in this paper. For each case, we compared the agent's final response, the structured answer extracted by the shared reference executor, and the gold answer under the predefined schema. We found no extraction errors in the audited cases (0/3933).

\section{Comparison between Real Apps and Simulators}
\begin{figure*}[!t]
    \centering
    \includegraphics[width=\textwidth]{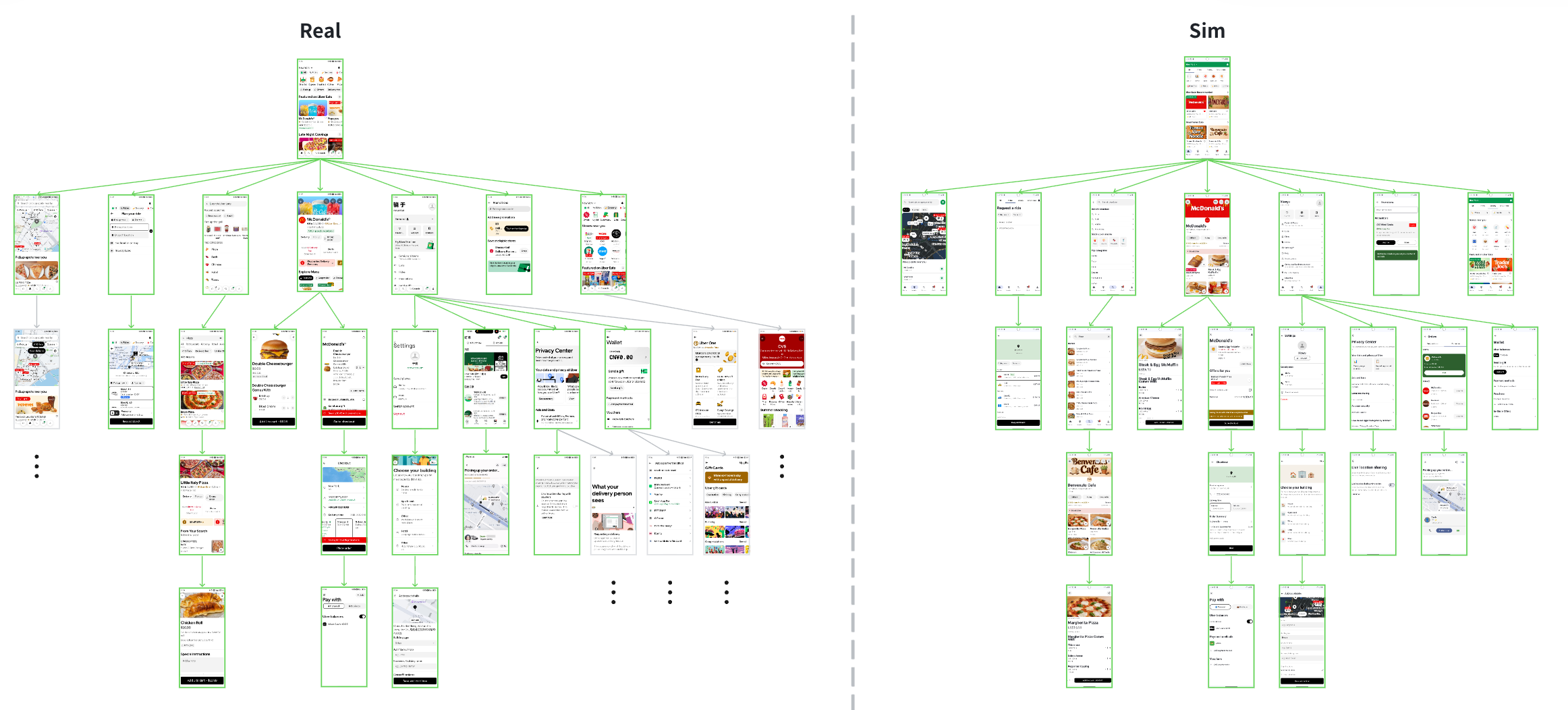}
    \caption{
    Page-transition comparison between a real app and its simulator.
    Green boxes indicate matched task-relevant pages in both apps, while black boxes mark real-app pages not retained in the simulator.
    }
    \label{fig_PageTree}
\end{figure*}

Figure~\ref{fig_PageTree} shows a page-transition comparison between a real app and its simulator. The simulator preserves the matched task-relevant pages and transitions, while excluding unmatched real-app states that are not required for the benchmark tasks.

\section{Benchmark Construction Quality Control}\label{app:construction-qc}

Each simulated app was implemented by a single lead developer and then independently inspected by several quality-control reviewers, whose reported issues were fixed before the app entered the benchmark. Reviewers executed both ``correct completion'' and ``deliberate failure'' trajectories and checked the True/False verdict of the evaluator case by case; no misjudged case was found.

To make the iterative nature of this process concrete, we recorded the full construction log of one app. We selected Uber Eats because it is the English app with the widest functional span in AppSim-Bench, covering food ordering, grocery, ride requests, map-based store selection, and cart management, so its pages exercise most of the interaction patterns encountered elsewhere. Table~\ref{tab:construction-rounds} reports, per page, the number of human review--correction rounds and the wall-clock construction time. No page was accepted on the first attempt: each required about 2.3 rounds on average, and the corrections span both visual presentation and interaction correctness.

\begin{table}[t]
\centering
\scriptsize
\caption{Human review--correction rounds during the construction of the Uber Eats simulator. ``Rounds'' counts the iterations of human inspection and correction before a page was accepted.}
\label{tab:construction-rounds}
\begin{tabular}{lrl}
\toprule
Page & Rounds & Representative correction \\
\midrule
Home & 3 & Realigned top icons, added real images \\
Cart & 3 & Split action buttons, linked store page \\
Profile & 2 & Unified icon style \\
Trips & 1 & Added ride-request form \\
Grocery & 2 & Added real images \\
Location & 2 & Made stores clickable, live filter \\
Search & 2 & Live input, clear, and results \\
Store Home & 2 & Real store data, dish-card entry \\
Product Detail & 3 & Add-to-cart return path \\
View Cart & 3 & Corrected quantities and images \\
\midrule
Total & 23 & $\approx$2.3 rounds per page, 2h22m \\
\bottomrule
\end{tabular}
\end{table}

\section{Behavioral Fidelity of Task-aware Pruning}\label{app:placeholder-residual}

AppSim-Bench reproduces the interactive elements and navigation logic of the real app on every retained page, and implements multiple valid paths per task. Fully task-irrelevant pages are not implemented; they render an explicit ``page under development'' placeholder. Such a placeholder is the one behavioral element that differs from a real app, since it signals early that the agent has left the task scope, whereas a real off-track page might let it drift further.

We therefore quantify the residual effect of pruning as the fraction of executions that entered a placeholder and still succeeded. This is a deliberately pessimistic upper bound: it credits every such success entirely to the placeholder, even though the agent may have recovered on its own.

Because this measure requires manually replaying trajectories to determine whether a placeholder was reached, we audited a subset rather than all 19 agents. We selected six apps spanning both ecosystems and four categories (video, food delivery, social, e-commerce, music), and three agents chosen to bracket the accuracy range observed in Table~\ref{tab:SR_overall_length_subtask_merged}: Claude-Opus-4.7 (highest), GPT-5 (intermediate), and Gemini-2.5-Pro (lowest among general-purpose agents). If the residual were driven by agent capability, this spread would expose it.

\begin{table}[t]
\centering
\scriptsize
\caption{Tasks that entered a ``page under development'' placeholder and still succeeded, reported as a pessimistic upper bound on the effect of task-aware pruning. Percentages are over each app's tasks.}
\label{tab:placeholder-residual}
\setlength{\tabcolsep}{3.5pt}
\begin{tabular}{lrrrr}
\toprule
App & Tasks & Claude-Opus & GPT-5 & Gemini-2.5 \\
\midrule
Bilibili & 30 & 2 (6.7\%) & 2 (6.7\%) & 1 (3.3\%) \\
Uber Eats & 34 & 1 (2.9\%) & 7 (20.6\%) & 1 (2.9\%) \\
Instagram & 40 & 1 (2.5\%) & 2 (5.0\%) & 1 (2.5\%) \\
NetEase Music & 39 & 2 (5.1\%) & 1 (2.6\%) & 0 (0.0\%) \\
JD & 40 & 2 (5.0\%) & 3 (7.5\%) & 0 (0.0\%) \\
YouTube & 35 & 0 (0.0\%) & 1 (2.9\%) & 1 (2.9\%) \\
\midrule
Total & 218 & 8 (3.67\%) & 16 (7.34\%) & 4 (1.83\%) \\
\bottomrule
\end{tabular}
\end{table}

Table~\ref{tab:placeholder-residual} shows that the residual is at most 7.34\% for any single agent and 4.28\% over all 654 executions, concentrated in a few app--agent combinations and near zero elsewhere. Replacing the placeholders with fully implemented pages would therefore shift the measured accuracies by at most a few points, and the pruned regions do not constitute an exploitable shortcut.

\section{Error Decomposition on Numerical-reasoning Tasks}\label{app:numerical-error-decomposition}

A numerical-reasoning task is executed in two stages: the agent first locates and reads the values that enter the computation from the interface (perception and extraction), then applies counting, arithmetic, comparison, or threshold filtering to them (arithmetic and logical reasoning). Failures can arise in either stage, and the two carry different implications for what the benchmark measures.

Separating them requires comparing, value by value, what the agent read against what the interface displayed. We did not automate this with a judge VLM, because a model prone to misreading cannot reliably decide whether another model misread. We therefore inspected screenshots manually, which bounds how much we could audit. We selected the numerical-reasoning tasks of JD as the audit target because it is the most numerically dense app in the benchmark (prices, discounts, quantities, ratings) and therefore places the heaviest load on perception; if extraction errors were common anywhere, they would appear here. We audited every execution step of the failed tasks for the same three agents used above.

\begin{table}[t]
\centering
\scriptsize
\caption{Manual audit of perception/extraction errors on the numerical-reasoning tasks of JD. Every execution step of each failed task was inspected.}
\label{tab:numerical-error-decomposition}
\setlength{\tabcolsep}{4pt}
\begin{tabular}{lrrrr}
\toprule
Agent & Failed & Steps & Err. & Rate \\
\midrule
Claude-Opus-4.7 & 17/27 & 284 & 0 & 0\% \\
GPT-5 & 20/27 & 443 & 0 & 0\% \\
Gemini-2.5-Pro & 25/27 & 560 & 1 & 0.18\% \\
\midrule
Total & --- & 1,287 & 1 & 0.08\% \\
\bottomrule
\end{tabular}
\end{table}

Across 1,287 execution steps, only one step exhibited a perception or extraction error. Almost all observed failures therefore arise in the reasoning stage rather than the perception stage. Notably, the agent with the most failed tasks (Gemini-2.5-Pro, 25/27) also shows a near-zero extraction rate, which indicates that the discriminating factor is reasoning rather than reading. This holds for the state-of-the-art agents audited here; for weaker agents, perception errors may become more frequent.

\section{Failure Modes of Unsolved Tasks}\label{app:unsolved-failure-modes}

We examined the 159 tasks that no evaluated agent solved. Their difficulty concentrates in three capability gaps. The first is \textbf{long-horizon multi-step workflows}, where the agent must track completed actions, interpret the current screen against the original instruction, and suppress the accumulation of intermediate errors. The second is \textbf{incomplete information gathering}, where the agent stops exploring too early, for example by inspecting only the visible portion of a list, and then decides on partial information. The third is \textbf{numerical reasoning chains on the interface}, where the agent must enumerate the relevant values before comparing or aggregating them, so that a break at any link causes failure.

The hardest tasks typically stack several of these gaps. In AMAZON\#21, the agent must find the three most expensive items in a long shopping cart, sum their prices, and return the total; the deterministic verifier expects 3118.98. No agent produced the correct value, and the outputs clustered at lower totals such as 1698.98, 1848.98, and 2868.98. Every reported total falls below the correct one, which is the signature of deciding before the list was fully enumerated. This case combines incomplete information gathering with a broken numerical-reasoning chain.

\section{App Info in AppSim-Bench}
\subsection{Overview of the Apps in AppSim-Bench}
We present the list of applications (Apps) included in AppSim-Bench in \Cref{tab:app_overview}.
These Apps span a diverse range of categories, including video sharing, travel services, food delivery, navigation, e-commerce, music streaming, social media, communication, ride hailing, and video conferencing, ensuring a comprehensive evaluation environment.
\begin{table*}[!t]
  \centering
  \begin{CJK*}{UTF8}{gbsn}
  \scriptsize
  \setlength{\tabcolsep}{1.5pt}
  \caption{Overview of Apps in AppSim-Bench. This table lists the Chinese and English names, a brief description, and the number of tasks for each App.} \label{tab:app_overview}
  \begin{tabular}{@{}>{\centering\arraybackslash}p{0.03\textwidth} p{0.095\textwidth} p{0.12\textwidth} p{0.54\textwidth} >{\centering\arraybackslash}p{0.06\textwidth}@{}}
    \toprule
    \textbf{ID} & \textbf{App Name (ZH)} & \textbf{App Name (EN)} & \textbf{Description} & \textbf{\#Tasks} \\
    \midrule
    1 & 哔哩哔哩 & Bilibili & A video-sharing app with ACG-oriented community content. & 30 \\
    2 & \textbackslash{} & Booking.com & A travel booking app for hotels, flights, cars, taxis, and attractions. & 25 \\
    3 & 携程旅行 & Ctrip & A travel-service app for hotels, flights, trains, trips, and messages. & 35 \\
    4 & 饿了么 & Ele.me & A local-life and food-delivery app for restaurants, orders, coupons, and addresses. & 40 \\
    5 & 高德地图 & Amap & A map and navigation app with nearby places, routes, favorites, and travel tools. & 28 \\
    6 & 网易云音乐 & NetEase Cloud Music & A music-streaming app with songs, playlists, charts, lyrics, and user profiles. & 39 \\
    7 & 京东 & JD & An e-commerce app for product search, carts, orders, addresses, and store services. & 40 \\
    8 & 小红书 & RedNote & A lifestyle social-media app for notes, creators, comments, messages, and profiles. & 29 \\
    9 & 腾讯会议 & Tencent Meeting & A video-conferencing app for meetings, contacts, messages, and personal rooms. & 26 \\
    10 & \textbackslash{} & Uber Eats & A food-delivery and local-services app with restaurants, carts, orders, and rides. & 34 \\
    11 & 微信 & WeChat & A messaging and social app with chats, contacts, groups, and Moments. & 16 \\
    12 & \textbackslash{} & YouTube & A video platform for search, playback, comments, subscriptions, Shorts, and settings. & 35 \\
    13 & \textbackslash{} & Amazon & An e-commerce app for product discovery, shopping lists, carts, orders, and support. & 40 \\
    14 & \textbackslash{} & WhatsApp & A messaging app with chats, calls, statuses, channels, communities, and contacts. & 40 \\
    15 & \textbackslash{} & Zoom & A video-meeting app for instant meetings, scheduled meetings, chat, contacts, and settings. & 20 \\
    16 & \textbackslash{} & Spotify & A music and podcast app with playback, search, playlists, lyrics, and subscriptions. & 40 \\
    17 & \textbackslash{} & Instagram & A social-media app for posts, reels, messages, profiles, search, and publishing. & 40 \\
    \bottomrule
  \end{tabular}
  \end{CJK*}
\end{table*}

\subsection{Screenshots of the Apps in AppSim-Bench}
In this section, we present screenshots for all 17 apps included in AppSim-Bench.
As shown in Figures~\ref{fig:app_screenshots_1}--\ref{fig:app_screenshots_3}, these screenshots provide an intuitive overview of the interface designs and functional layouts covered by the benchmark.

\newcommand{\appsimshot}[2]{%
    \begin{subfigure}[t]{0.31\linewidth}
        \centering
        \includegraphics[height=0.29\textheight,keepaspectratio]{#1}
        \caption{#2}
    \end{subfigure}%
}

\begin{figure*}[p]
    \centering
    \appsimshot{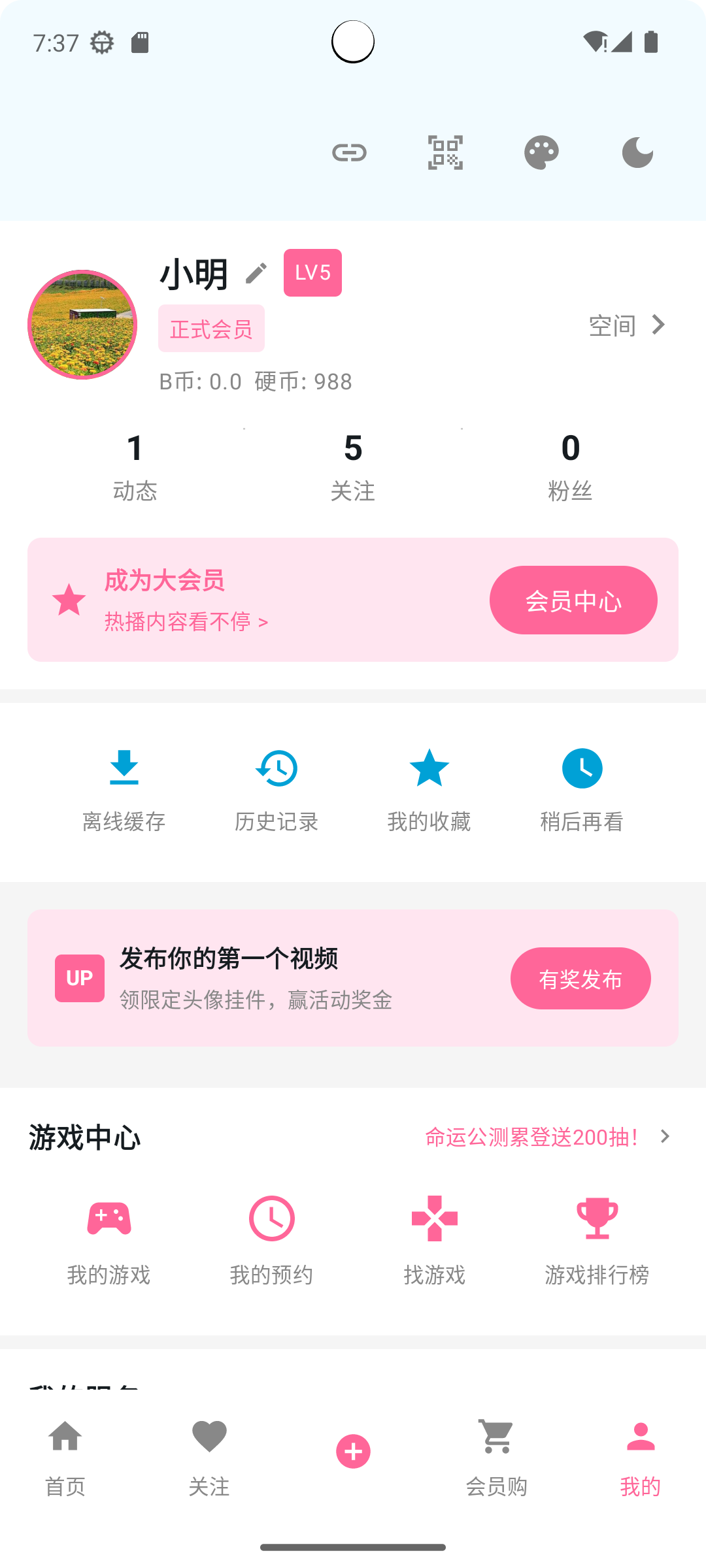}{Bilibili}
    \hfill
    \appsimshot{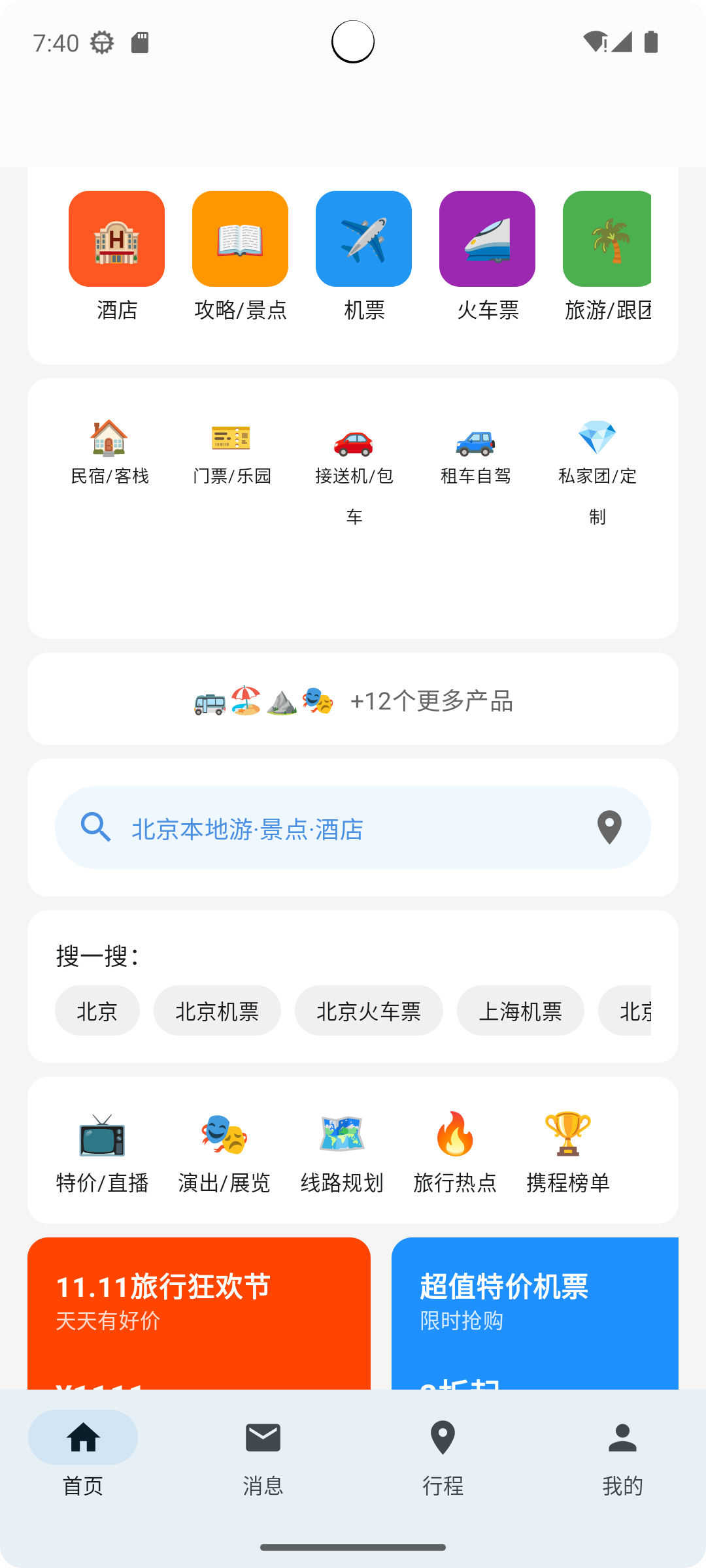}{Ctrip}
    \hfill
    \appsimshot{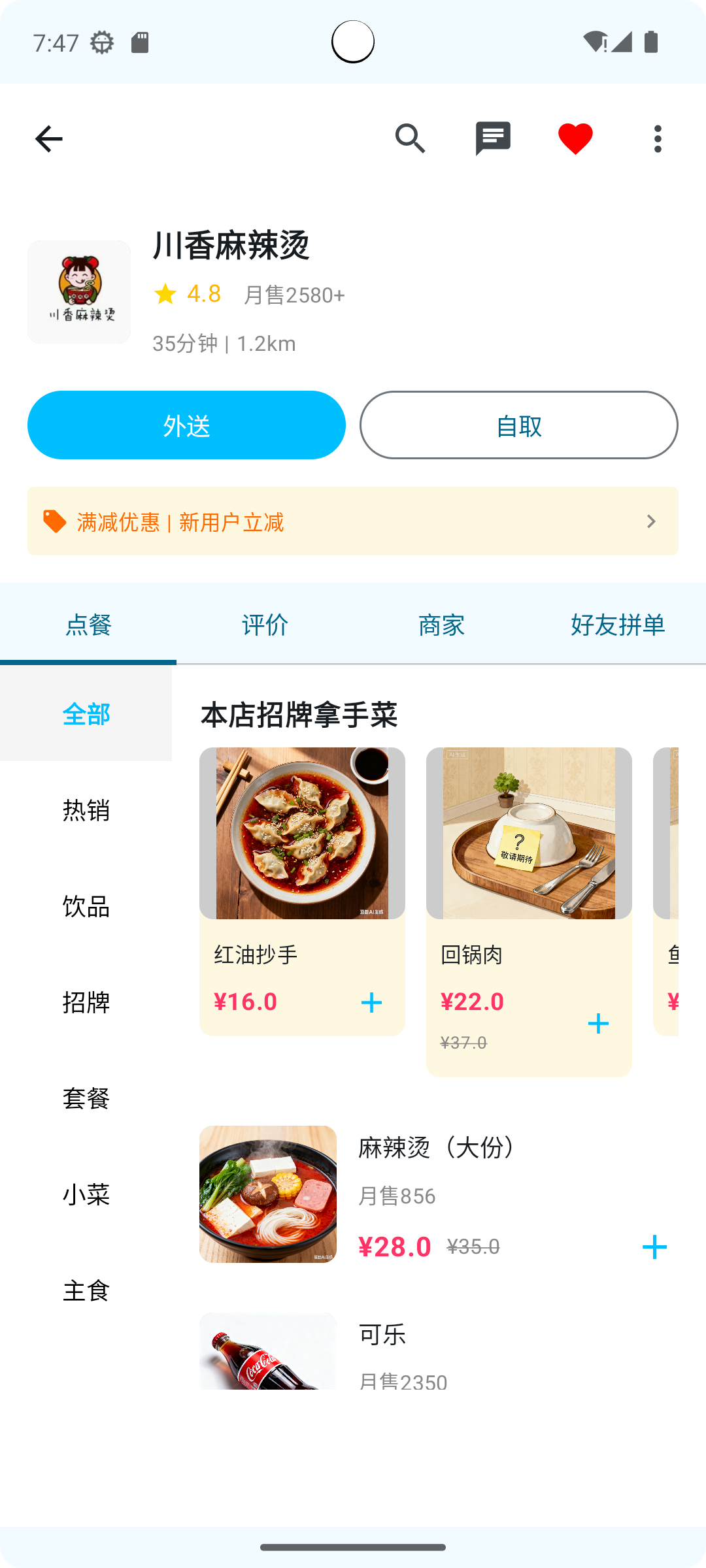}{Ele.me}

    \vspace{0.6em}
    \appsimshot{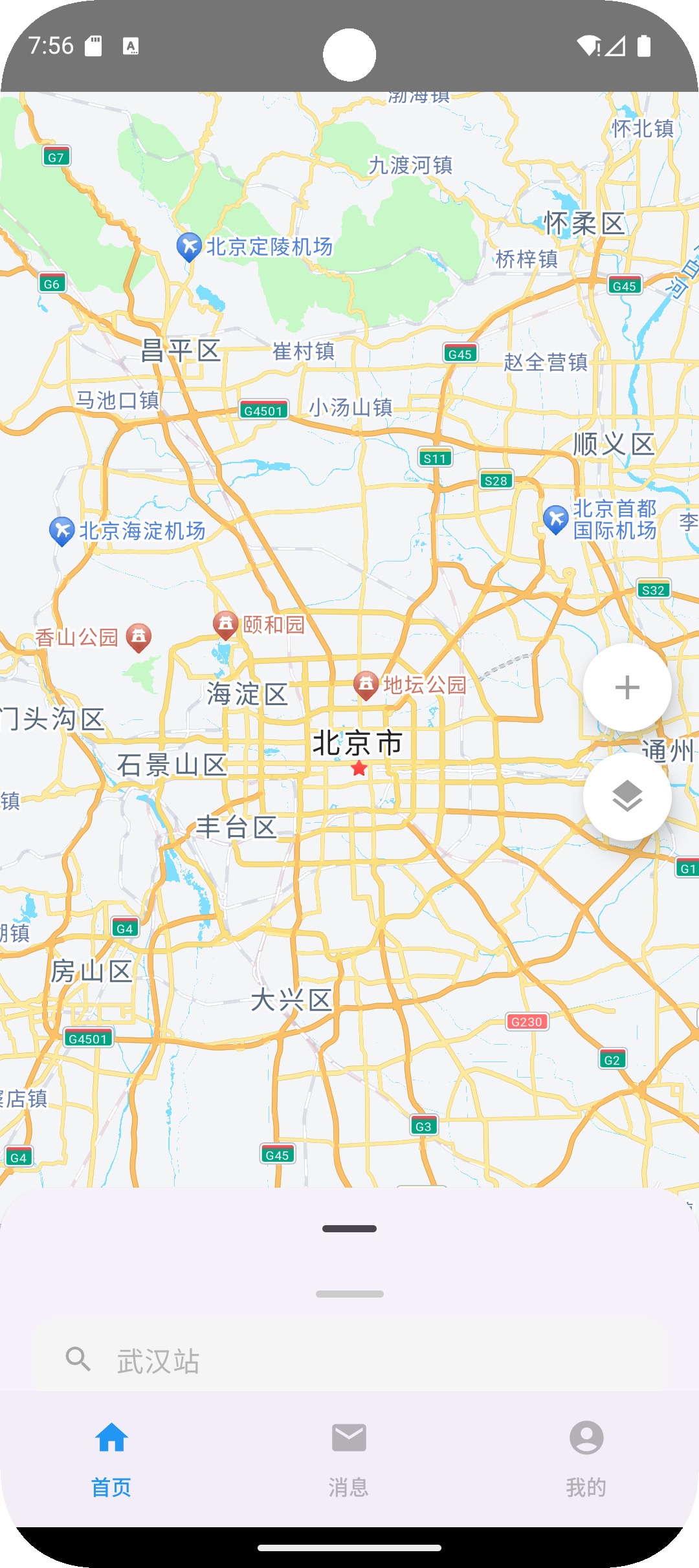}{Amap}
    \hfill
    \appsimshot{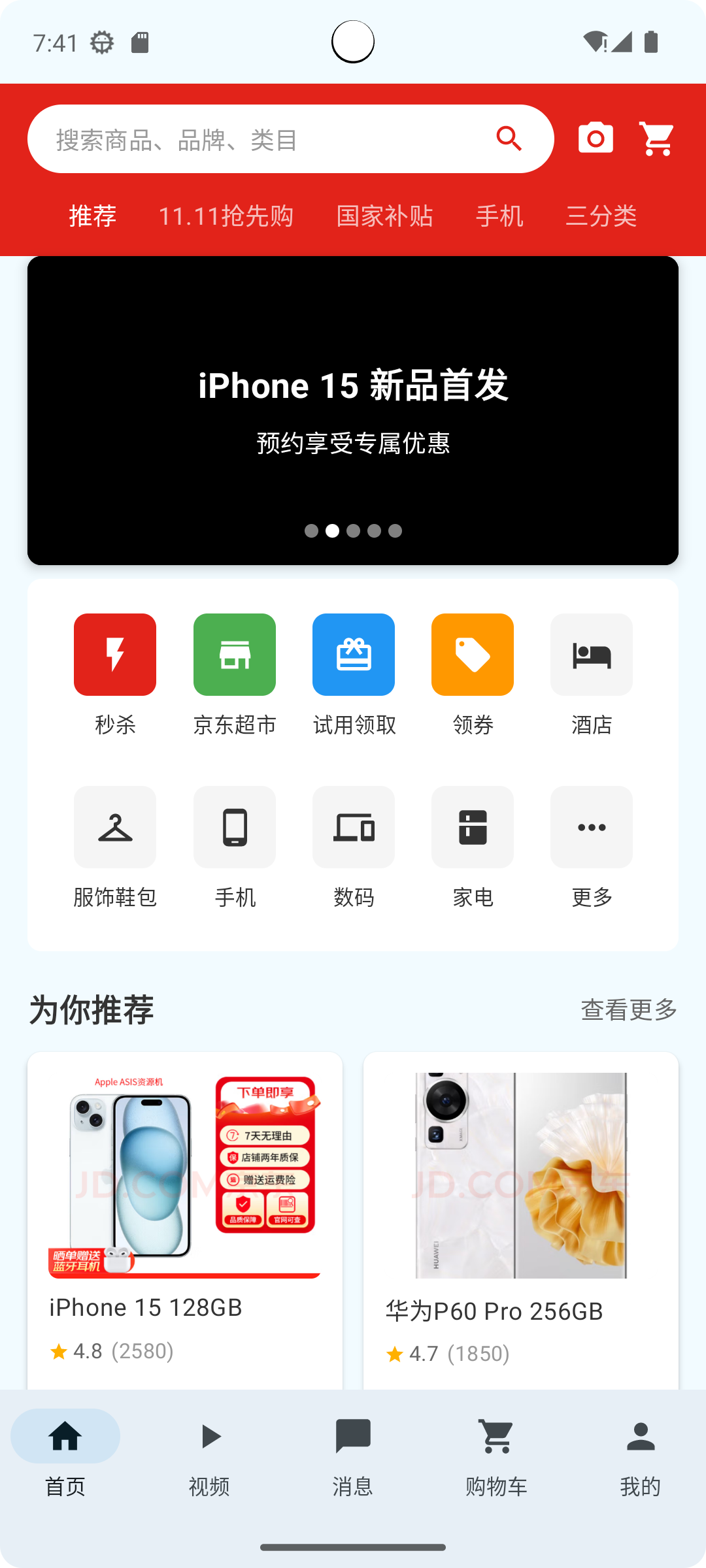}{JD}
    \hfill
    \appsimshot{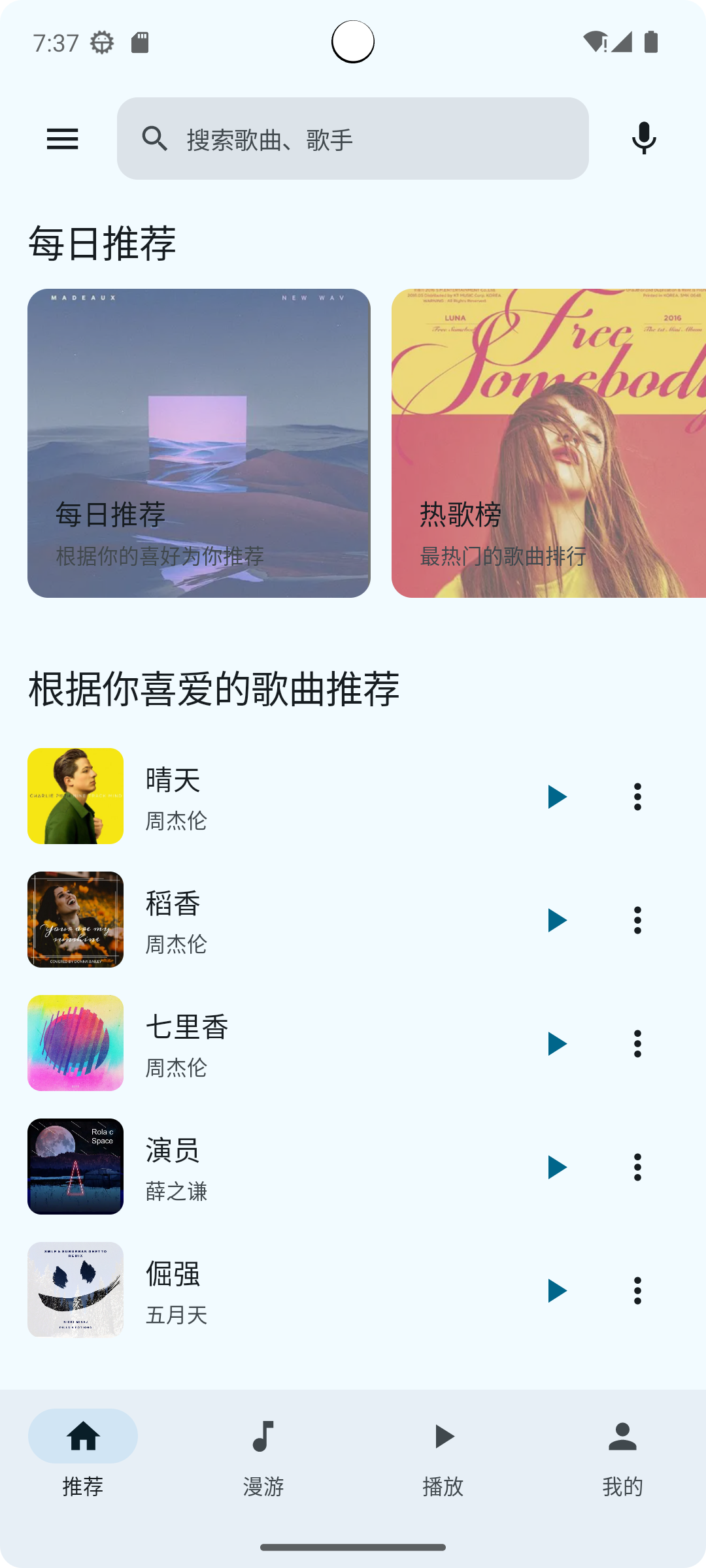}{NetEase Cloud Music}
    \caption{Screenshots of the Apps in AppSim-Bench (1/3).}
    \label{fig:app_screenshots_1}
\end{figure*}

\begin{figure*}[p]
    \centering
    \appsimshot{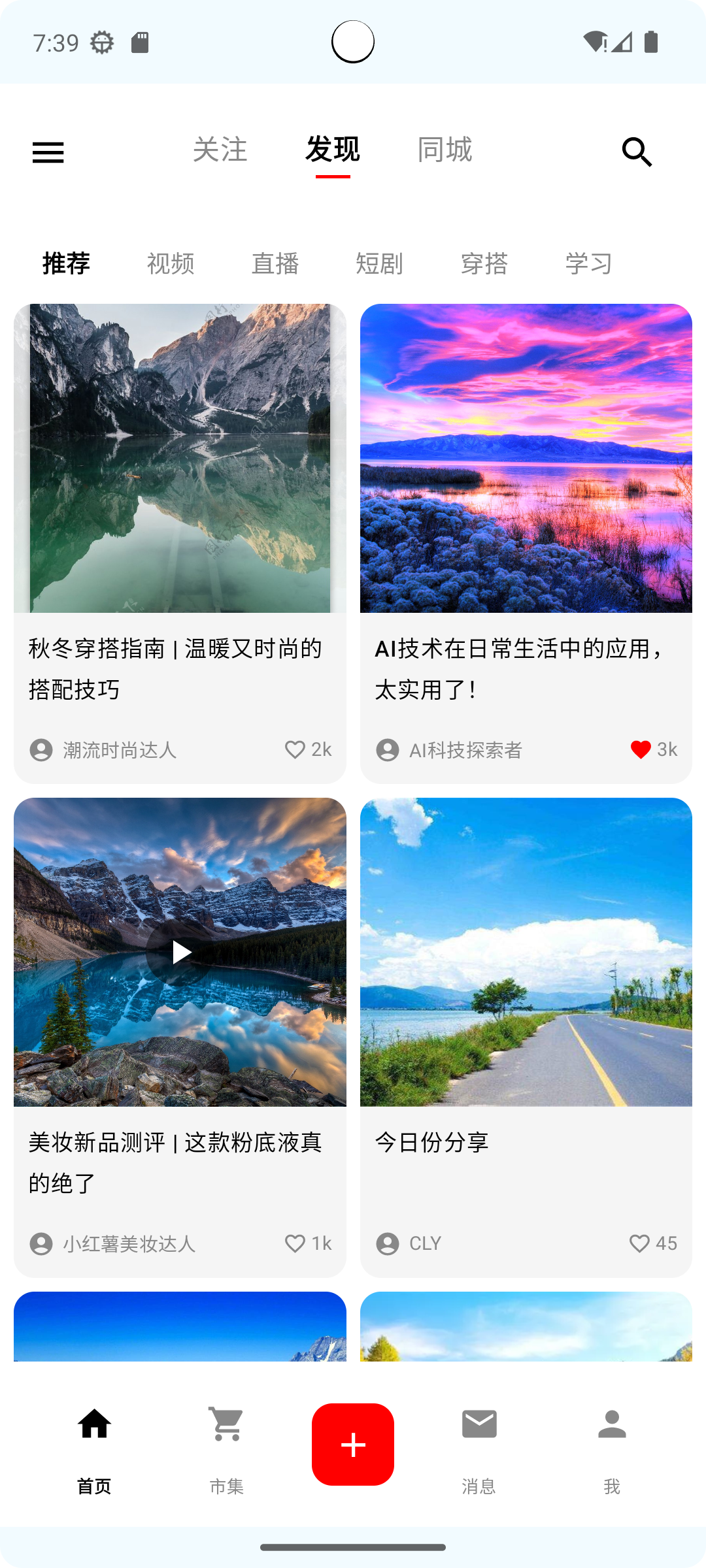}{RedNote}
    \hfill
    \appsimshot{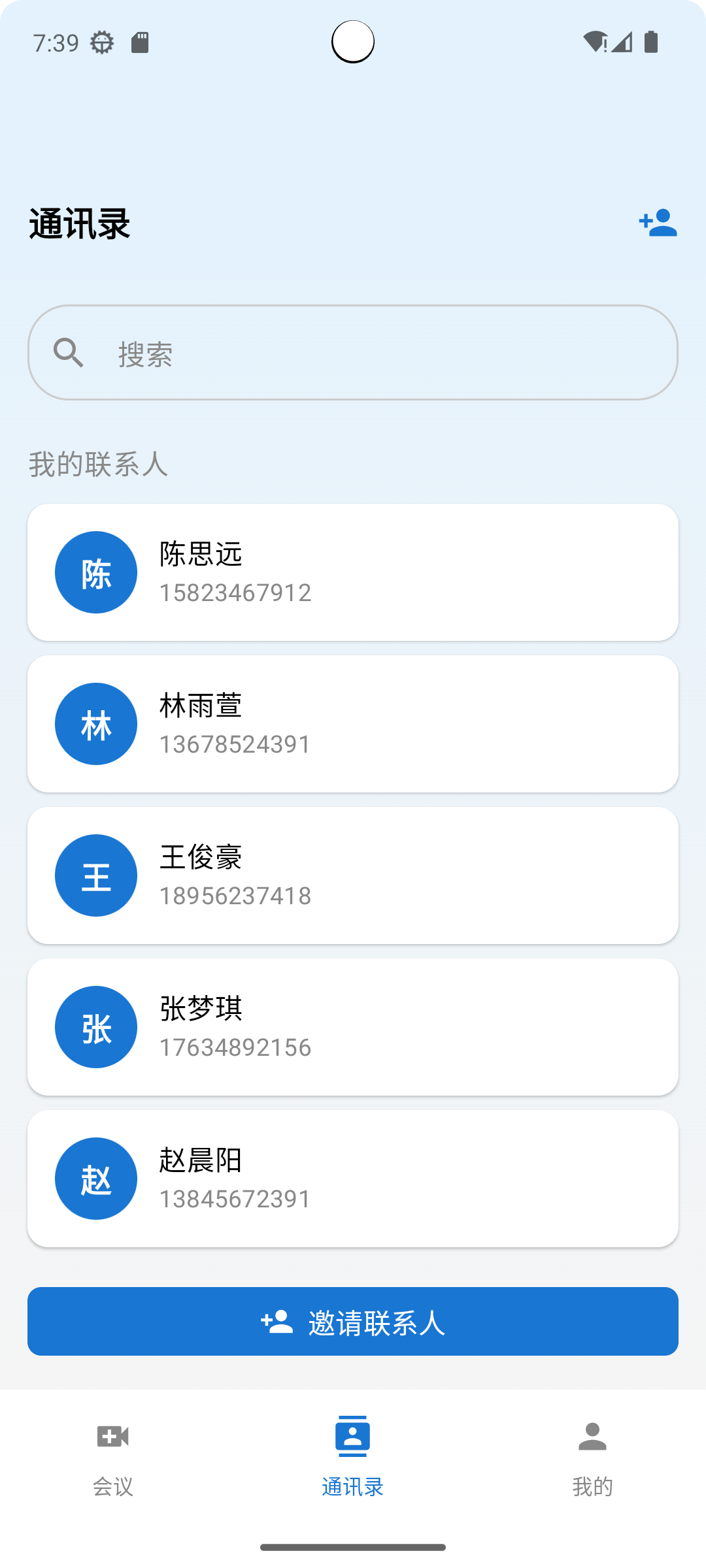}{Tencent Meeting}
    \hfill
    \appsimshot{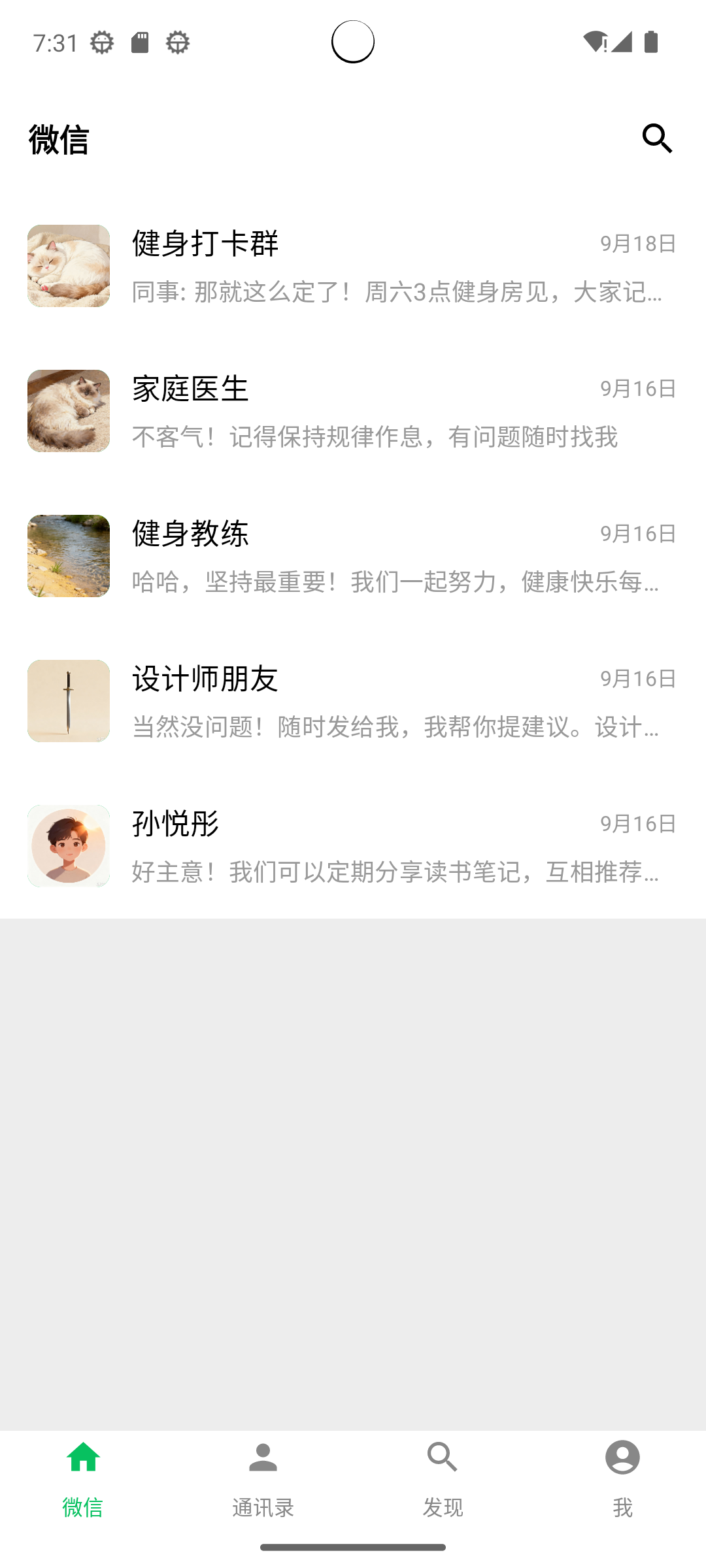}{WeChat}

    \vspace{0.6em}
    \appsimshot{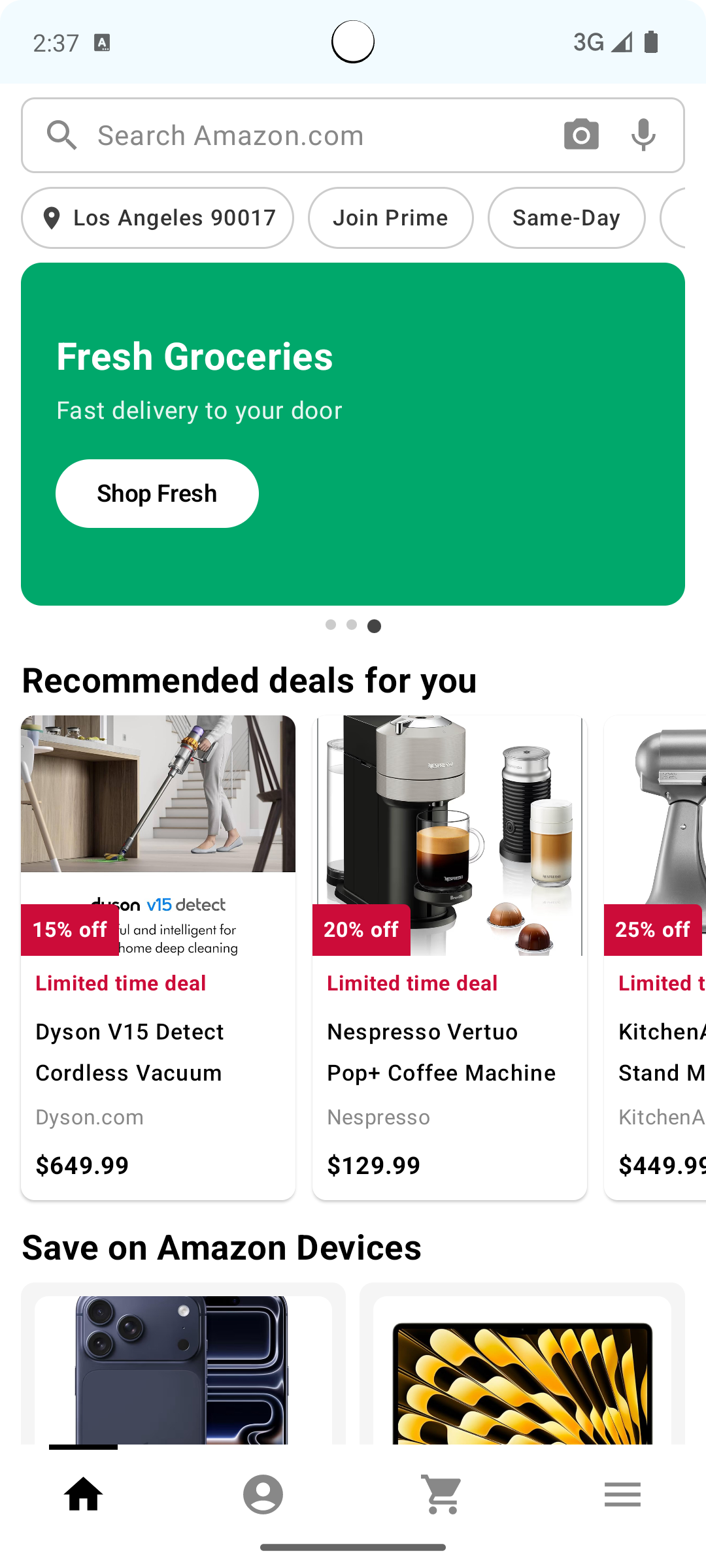}{Amazon}
    \hfill
    \appsimshot{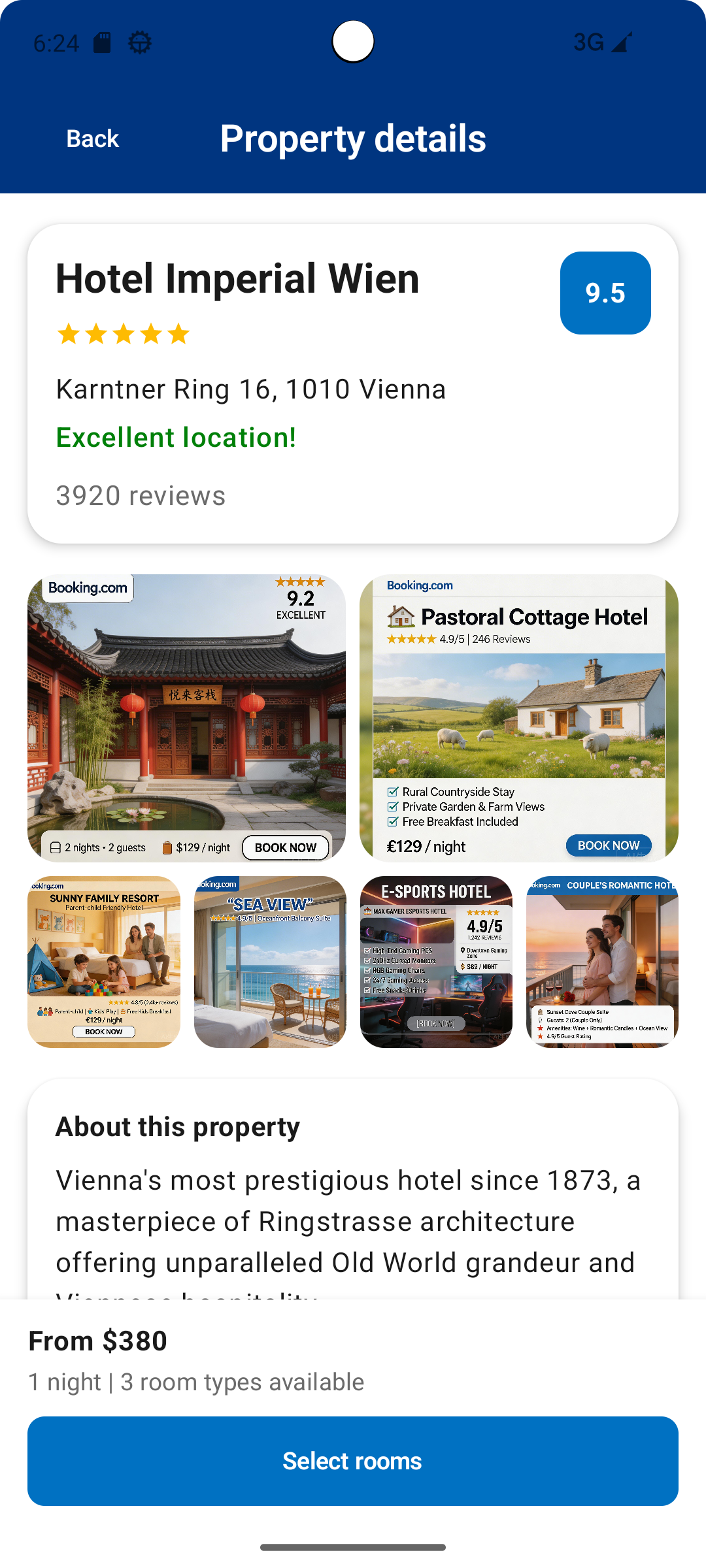}{Booking.com}
    \hfill
    \appsimshot{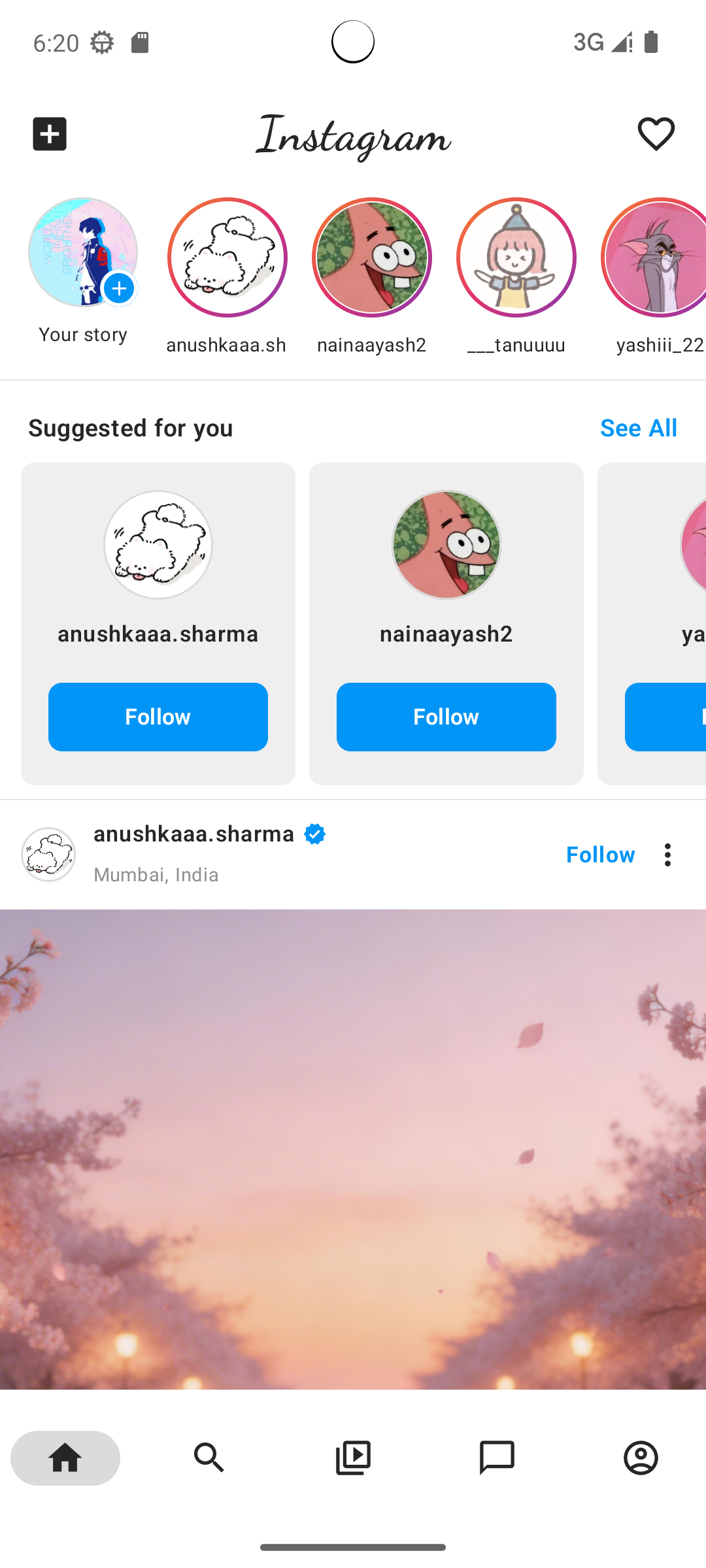}{Instagram}
    \caption{Screenshots of the Apps in AppSim-Bench (2/3).}
    \label{fig:app_screenshots_2}
\end{figure*}

\begin{figure*}[p]
    \centering
    \appsimshot{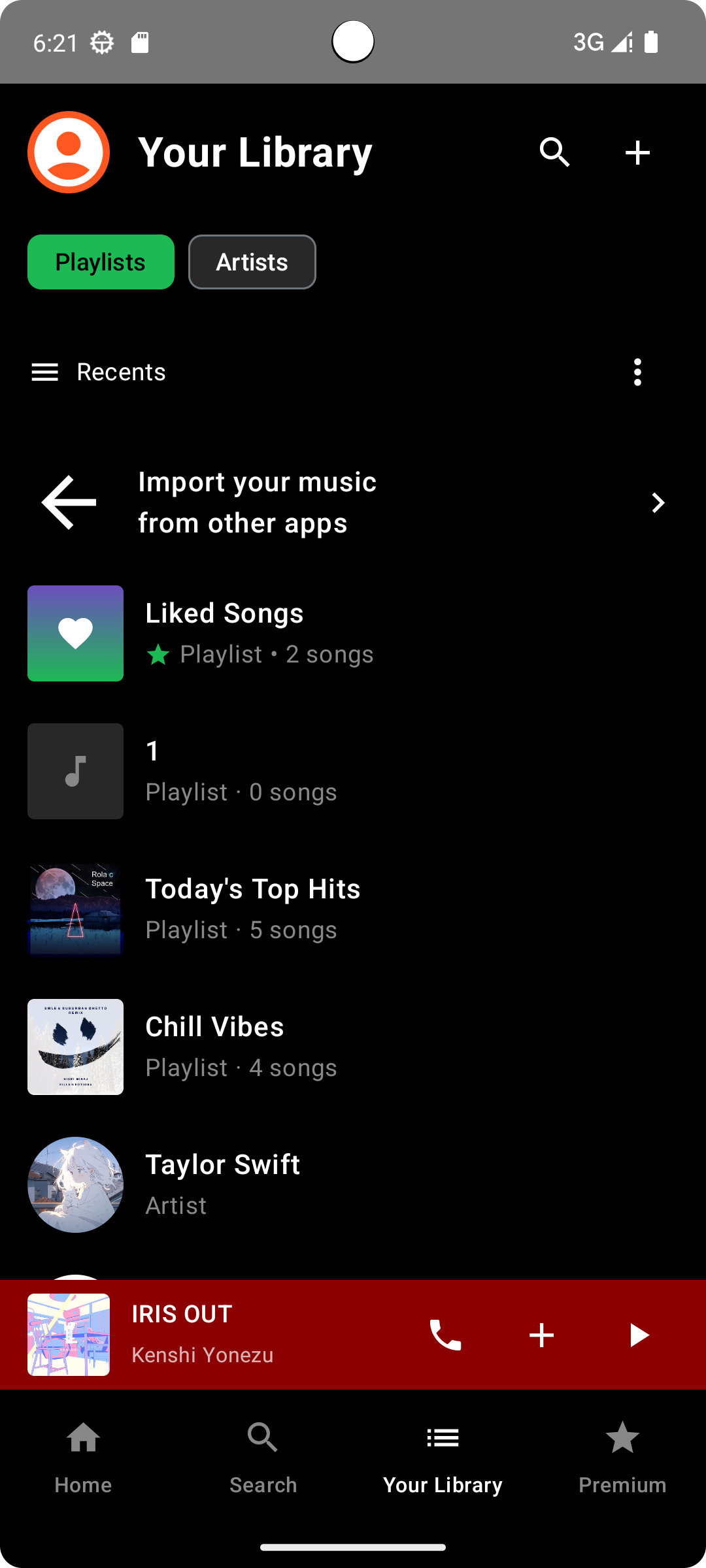}{Spotify}
    \hfill
    \appsimshot{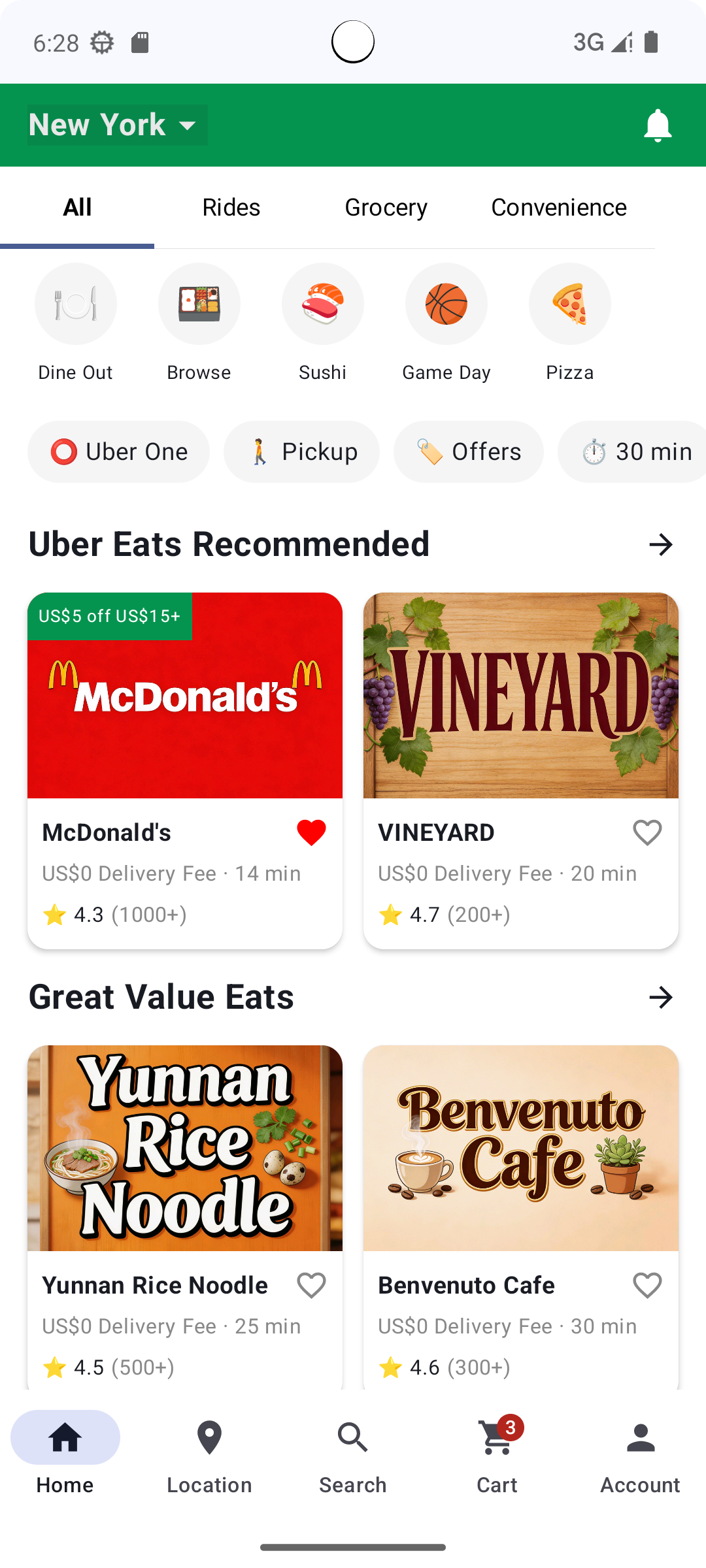}{Uber Eats}
    \hfill
    \appsimshot{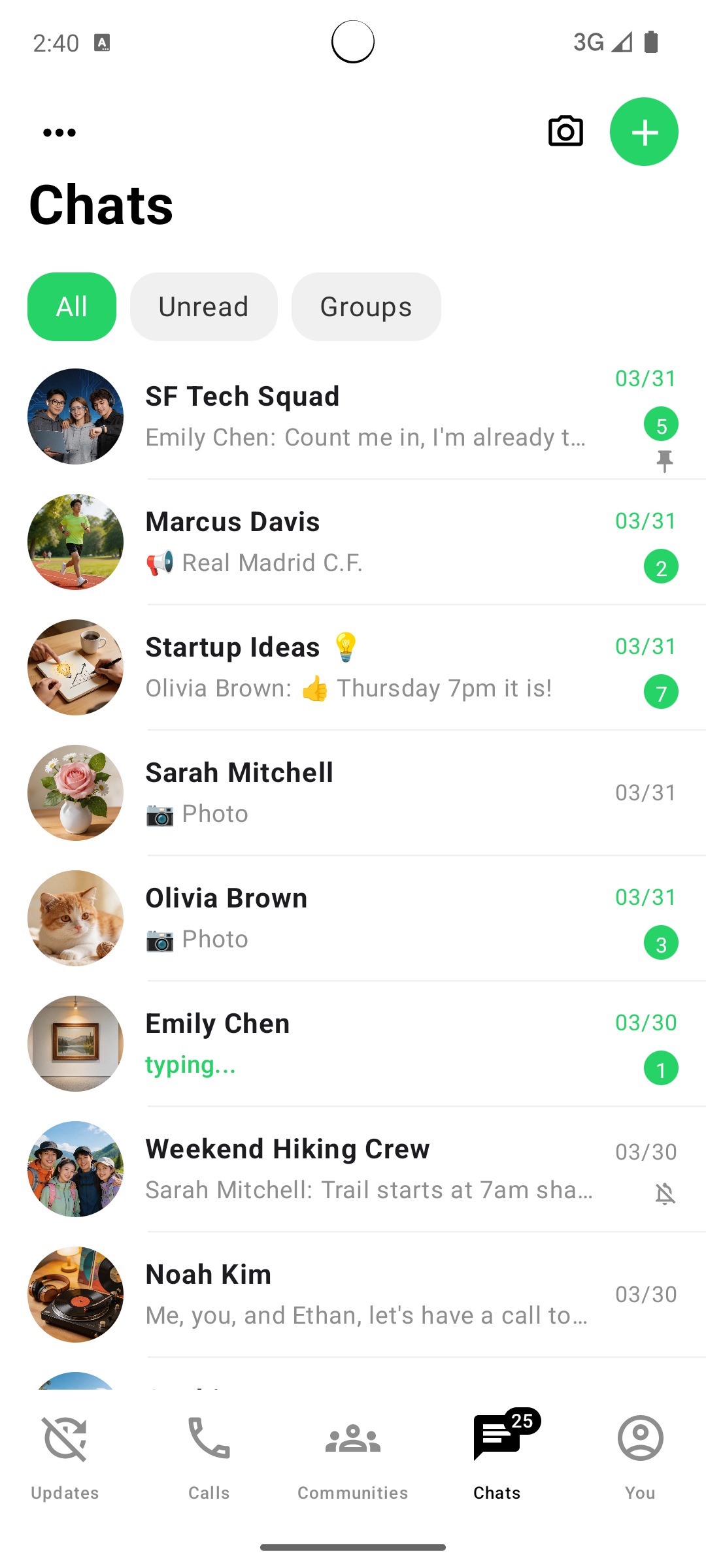}{WhatsApp}

    \vspace{0.6em}
    \appsimshot{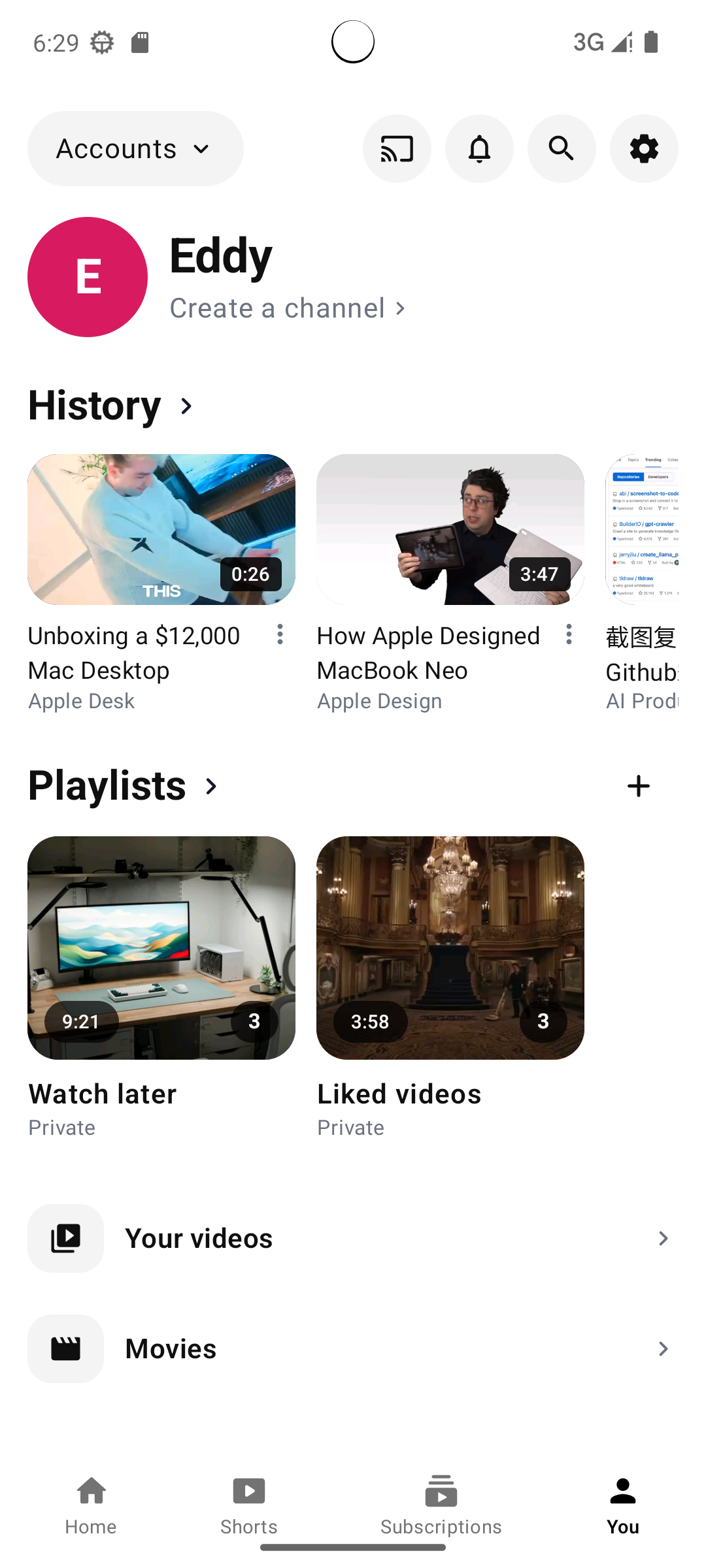}{YouTube}
    \hfill
    \appsimshot{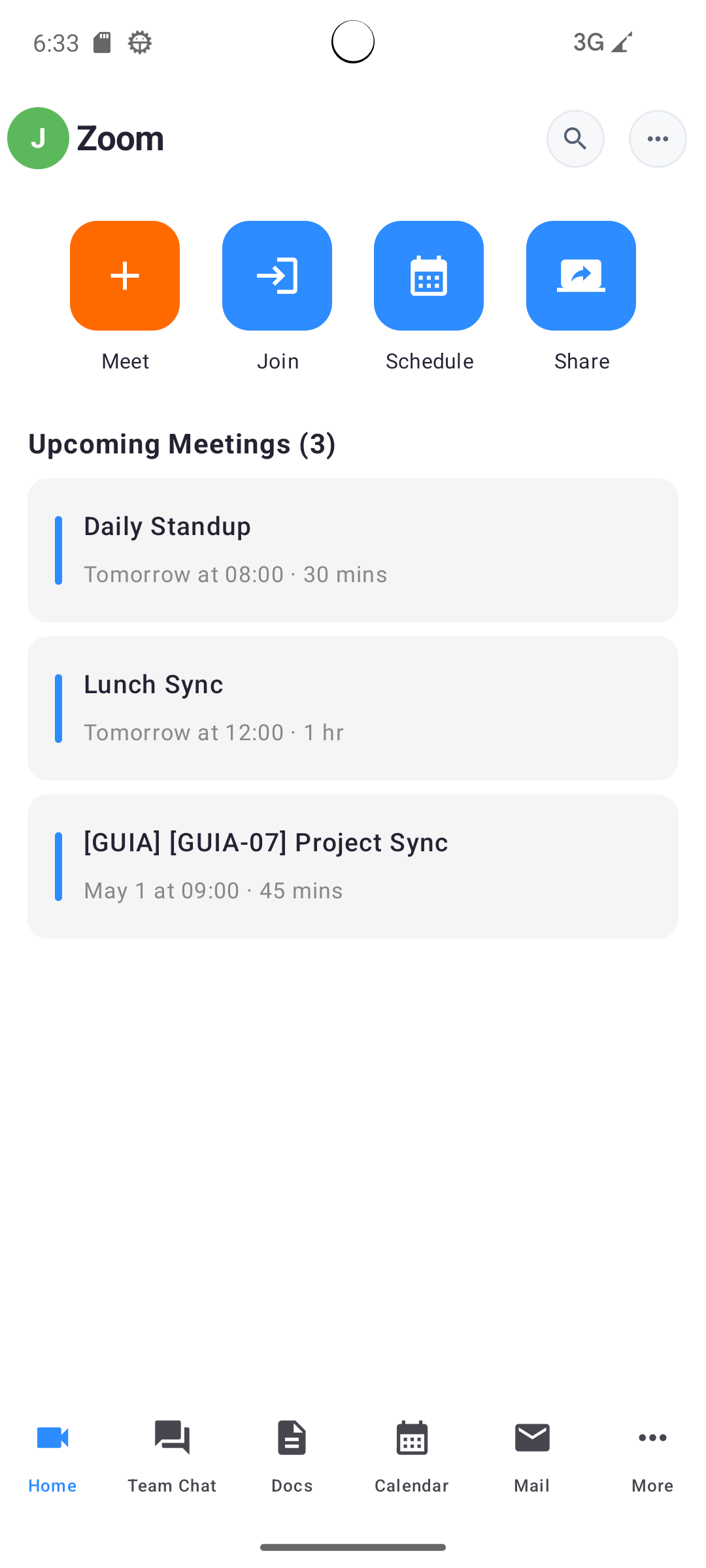}{Zoom}
    \hfill
    \begin{subfigure}[t]{0.31\linewidth}
        \centering
        \vspace{0.29\textheight}
        \caption*{}
    \end{subfigure}
    \caption{Screenshots of the Apps in AppSim-Bench (3/3).}
    \label{fig:app_screenshots_3}
\end{figure*}

\subsection{Statistics of AppSim-Bench} \label{sec:app_sim}
Table~\ref{tab:current-task-counts} reports the complete app-level task statistics of AppSim-Bench, including Chinese/English app membership, task-length buckets, and numerical-reasoning subtype counts.

\begin{table*}[t]
\centering
\begingroup
\newcommand{\yesmark}{\textcolor{green!60!black}{\checkmark}}
\newcommand{\nomark}{\textcolor{red!80!black}{$\times$}}
\caption{Task statistics of AppSim-Bench by app. ZH and EN indicate whether an app belongs to the Chinese or English app group. The task-length buckets are based on the human-annotated reference steps $\#S$ required to complete a task. ``Cnt.'', ``Arith.'', ``Comp.'', and ``Thres.'' denote counting, arithmetic calculation, numeric comparison, and threshold filtering, respectively. Numerical-reasoning subtypes are not mutually exclusive.}
\label{tab:current-task-counts}
\resizebox{\textwidth}{!}{%
\begin{tabular}{lrccrrrrrrrr}
\toprule
\multirow{2}{*}{App} &
\multicolumn{3}{c}{Overall Benchmark} &
\multicolumn{3}{c}{Task Length} &
\multicolumn{5}{c}{Numerical Reasoning Tasks} \\
\cmidrule(lr){2-4}\cmidrule(lr){5-7}\cmidrule(lr){8-12}
& Total & ZH & EN
& $\#S_{\leq 5}$ & $\#S_{6\text{-}10}$ & $\#S_{\geq 11}$
& Total & Cnt. & Arith. & Comp. & Thres. \\
\midrule
Amazon & 40 & \nomark & \yesmark & 10 & 15 & 15 & 21 & 5 & 4 & 14 & 2 \\
Bilibili & 30 & \yesmark & \nomark & 20 & 5 & 5 & 17 & 4 & 8 & 6 & 0 \\
Booking & 25 & \nomark & \yesmark & 5 & 8 & 12 & 9 & 0 & 1 & 6 & 2 \\
Ctrip & 35 & \yesmark & \nomark & 15 & 8 & 12 & 8 & 0 & 3 & 6 & 4 \\
Ele.me & 40 & \yesmark & \nomark & 25 & 9 & 6 & 21 & 8 & 6 & 9 & 2 \\
Gaode & 28 & \yesmark & \nomark & 14 & 9 & 5 & 11 & 1 & 1 & 8 & 2 \\
Instagram & 40 & \nomark & \yesmark & 30 & 5 & 5 & 2 & 2 & 0 & 0 & 0 \\
JD & 40 & \yesmark & \nomark & 10 & 20 & 10 & 27 & 7 & 6 & 12 & 6 \\
NetEase Cloud Music & 39 & \yesmark & \nomark & 26 & 8 & 5 & 3 & 2 & 0 & 1 & 0 \\
RedNote & 29 & \yesmark & \nomark & 17 & 7 & 5 & 6 & 4 & 1 & 1 & 0 \\
Spotify & 40 & \nomark & \yesmark & 30 & 5 & 5 & 3 & 3 & 0 & 0 & 1 \\
Tencent Meeting & 26 & \yesmark & \nomark & 15 & 5 & 6 & 10 & 8 & 1 & 1 & 0 \\
Uber Eats & 34 & \nomark & \yesmark & 19 & 6 & 9 & 15 & 2 & 9 & 5 & 0 \\
WeChat & 16 & \yesmark & \nomark & 6 & 5 & 5 & 5 & 5 & 0 & 0 & 0 \\
WhatsApp & 40 & \nomark & \yesmark & 10 & 15 & 15 & 7 & 3 & 0 & 3 & 1 \\
YouTube & 35 & \nomark & \yesmark & 25 & 5 & 5 & 13 & 8 & 5 & 0 & 0 \\
Zoom & 20 & \nomark & \yesmark & 5 & 5 & 10 & 5 & 3 & 0 & 3 & 1 \\
\midrule
ZH Apps & 283 & \yesmark & \nomark & 148 & 76 & 59 & 108 & 39 & 26 & 44 & 14 \\
EN Apps & 274 & \nomark & \yesmark & 134 & 64 & 76 & 75 & 26 & 19 & 31 & 7 \\
All Apps & 557 & \yesmark & \yesmark & 282 & 140 & 135 & 183 & 65 & 45 & 75 & 21 \\
\bottomrule
\end{tabular}%
}
\endgroup
\end{table*}

\section{More Experimental Results} \label{sec:app_exp} \label{app:app-level-accuracy}
Tables~\ref{tab:app-acc-zh} and~\ref{tab:app-acc-en} report the complete app-level accuracy results for Chinese and English apps, respectively.
The compact main-text summary in Table~\ref{tab:overall-acc-summary} is derived from these tables.
\begin{table*}[t]
\centering
\caption{Accuracy on Chinese apps. The first numeric column reports task-weighted accuracy over the Chinese app group. NetEase Music denotes the NetEase Cloud Music simulation. \textbf{Bold} and \underline{underlining} indicate the best and second-best values in each column.}
\label{tab:app-acc-zh}
\resizebox{\textwidth}{!}{%
\begin{tabular}{lrrrrrrrrrr}
\toprule
Agent Model & ZH Apps & Bilibili & Ctrip & Ele.me & Gaode & JD & RedNote & Tencent Meeting & WeChat & NetEase Music \\
\midrule
\multicolumn{11}{c}{\textit{General-purpose Agents}} \\
\midrule
Gemini-3-Pro & \textbf{60.07\%} & \underline{80.00\%} & \underline{77.14\%} & 55.00\% & \underline{67.86\%} & \underline{55.00\%} & 24.14\% & \textbf{57.69\%} & \textbf{81.25\%} & 53.85\% \\
Claude-Opus-4.7 & \textbf{60.07\%} & 63.33\% & 65.71\% & \textbf{72.50\%} & \textbf{78.57\%} & 52.50\% & 27.59\% & \textbf{57.69\%} & \underline{68.75\%} & \underline{56.41\%} \\
Gemini-3.1-Pro & \underline{59.36\%} & 73.33\% & \textbf{80.00\%} & 55.00\% & 57.14\% & \textbf{60.00\%} & 24.14\% & \textbf{57.69\%} & \underline{68.75\%} & \textbf{58.97\%} \\
GPT-5.5 & 58.66\% & \textbf{90.00\%} & 71.43\% & \underline{65.00\%} & 53.57\% & \textbf{60.00\%} & 27.59\% & 46.15\% & 43.75\% & \underline{56.41\%} \\
Doubao-Seed-1.6 & 50.88\% & 56.67\% & 74.29\% & 55.00\% & 46.43\% & 32.50\% & 41.38\% & 65.38\% & 18.75\% & 53.85\% \\
Qwen-3.6-Plus & 39.58\% & 20.00\% & 62.86\% & 52.50\% & 64.29\% & 20.00\% & \textbf{37.93\%} & 38.46\% & 18.75\% & 33.33\% \\
Qwen-3.6-Flash & 38.87\% & 26.67\% & 60.00\% & 45.00\% & 53.57\% & 27.50\% & \underline{31.03\%} & \underline{53.85\%} & 6.25\% & 33.33\% \\
GPT-5 & 37.81\% & 46.67\% & 17.14\% & 37.50\% & 50.00\% & 32.50\% & 20.69\% & 42.31\% & 62.50\% & 46.15\% \\
Doubao-Seed-2.0 & 36.40\% & 43.33\% & 48.57\% & 45.00\% & 42.86\% & 25.00\% & \underline{31.03\%} & 42.31\% & 25.00\% & 23.08\% \\
GPT-5.4 & 34.63\% & 50.00\% & 62.86\% & 37.50\% & 25.00\% & 25.00\% & 17.24\% & 19.23\% & 18.75\% & 41.03\% \\
Doubao-Seed-1.8 & 34.63\% & 26.67\% & 57.14\% & 35.00\% & 50.00\% & 27.50\% & 20.69\% & 46.15\% & 6.25\% & 30.77\% \\
Claude-Sonnet-4.6 & 31.10\% & 50.00\% & 22.86\% & 45.00\% & 21.43\% & 30.00\% & 24.14\% & 30.77\% & 12.50\% & 30.77\% \\
Claude-Sonnet-4.5 & 24.03\% & 26.67\% & 22.86\% & 25.00\% & 14.29\% & 17.50\% & 10.34\% & 26.92\% & 31.25\% & 41.03\% \\
Claude-Haiku-4.5 & 22.61\% & 30.00\% & 45.71\% & 25.00\% & 10.71\% & 17.50\% & 13.79\% & 7.69\% & 0.00\% & 33.33\% \\
Gemini-2.5-Pro & 21.20\% & 36.67\% & 17.14\% & 37.50\% & 3.57\% & 10.00\% & 13.79\% & 23.08\% & 12.50\% & 28.21\% \\
\midrule
\multicolumn{11}{c}{\textit{GUI-specialized Agents}} \\
\midrule
MobileAgent-v3.5 & 35.34\% & 30.00\% & 68.57\% & 32.50\% & 50.00\% & 27.50\% & 20.69\% & 30.77\% & 12.50\% & 33.33\% \\
UI-TARS-1.5-7B & 31.10\% & 16.67\% & 51.43\% & 35.00\% & 53.57\% & 15.00\% & 27.59\% & 26.92\% & 12.50\% & 33.33\% \\
V-Droid & 13.43\% & 20.00\% & 25.71\% & 15.00\% & 0.00\% & 17.50\% & 3.45\% & 15.38\% & 6.25\% & 10.26\% \\
AgentCPM-GUI & 9.19\% & 0.00\% & 34.29\% & 5.00\% & 7.14\% & 5.00\% & 6.90\% & 7.69\% & 0.00\% & 10.26\% \\
\bottomrule
\end{tabular}%
}
\end{table*}

\begin{table*}[t]
\centering
\caption{Accuracy on English apps. The first numeric column reports task-weighted accuracy over the English app group. \textbf{Bold} and \underline{underlining} indicate the best and second-best values in each column.}
\label{tab:app-acc-en}
\resizebox{\textwidth}{!}{%
\begin{tabular}{lrrrrrrrrr}
\toprule
Agent Model & EN Apps & Amazon & Booking & Instagram & Spotify & UberEats & WhatsApp & YouTube & Zoom \\
\midrule
\multicolumn{10}{c}{\textit{General-purpose Agents}} \\
\midrule
GPT-5.5 & \textbf{40.51\%} & \textbf{55.00\%} & \underline{48.00\%} & \textbf{40.00\%} & \textbf{27.50\%} & \textbf{47.06\%} & \textbf{12.50\%} & \underline{57.14\%} & 45.00\% \\
Claude-Opus-4.7 & \underline{40.15\%} & 47.50\% & 44.00\% & \underline{37.50\%} & 22.50\% & \textbf{47.06\%} & \textbf{12.50\%} & \textbf{65.71\%} & 60.00\% \\
Gemini-3.1-Pro & 39.42\% & \underline{52.50\%} & \textbf{56.00\%} & 27.50\% & 22.50\% & \underline{41.18\%} & \underline{10.00\%} & \underline{57.14\%} & \textbf{75.00\%} \\
Gemini-3-Pro & 39.42\% & \underline{52.50\%} & \textbf{56.00\%} & 27.50\% & \underline{25.00\%} & \underline{41.18\%} & \textbf{12.50\%} & \underline{57.14\%} & \underline{65.00\%} \\
Doubao-Seed-1.8 & 29.56\% & 27.50\% & \underline{48.00\%} & 27.50\% & 20.00\% & \textbf{47.06\%} & 5.00\% & 42.86\% & 30.00\% \\
GPT-5 & 28.83\% & 45.00\% & 28.00\% & 35.00\% & 17.50\% & 29.41\% & 5.00\% & 37.14\% & 40.00\% \\
Doubao-Seed-1.6 & 27.01\% & 20.00\% & 32.00\% & 30.00\% & 17.50\% & 29.41\% & 5.00\% & 54.29\% & 40.00\% \\
GPT-5.4 & 23.36\% & 25.00\% & 36.00\% & 32.50\% & 15.00\% & 17.65\% & \underline{10.00\%} & 31.43\% & 25.00\% \\
Claude-Sonnet-4.6 & 23.36\% & 17.50\% & 36.00\% & 30.00\% & 15.00\% & 23.53\% & \underline{10.00\%} & 45.71\% & 10.00\% \\
Qwen-3.6-Plus & 22.99\% & 27.50\% & 40.00\% & 20.00\% & 12.50\% & 29.41\% & 7.50\% & 34.29\% & 20.00\% \\
Claude-Haiku-4.5 & 22.99\% & 32.50\% & 12.00\% & 30.00\% & 15.00\% & 20.59\% & 7.50\% & 45.71\% & 15.00\% \\
Doubao-Seed-2.0 & 21.90\% & 10.00\% & 36.00\% & 20.00\% & 22.50\% & 29.41\% & 7.50\% & 37.14\% & 20.00\% \\
Claude-Sonnet-4.5 & 17.88\% & 12.50\% & 24.00\% & 25.00\% & 20.00\% & 14.71\% & 7.50\% & 31.43\% & 5.00\% \\
Qwen-3.6-Flash & 17.15\% & 15.00\% & 12.00\% & 15.00\% & 15.00\% & 32.35\% & 5.00\% & 25.71\% & 20.00\% \\
Gemini-2.5-Pro & 13.14\% & 10.00\% & 12.00\% & 20.00\% & 17.50\% & 11.76\% & 5.00\% & 22.86\% & 0.00\% \\
\midrule
\multicolumn{10}{c}{\textit{GUI-specialized Agents}} \\
\midrule
MobileAgent-v3.5 & 27.37\% & 27.50\% & 52.00\% & 20.00\% & 17.50\% & 32.35\% & 7.50\% & 48.57\% & 25.00\% \\
UI-TARS-1.5-7B & 13.87\% & 7.50\% & 12.00\% & 12.50\% & 15.00\% & 17.65\% & 2.50\% & 28.57\% & 20.00\% \\
AgentCPM-GUI & 10.95\% & 5.00\% & 4.00\% & 22.50\% & 12.50\% & 11.76\% & 2.50\% & 20.00\% & 5.00\% \\
V-Droid & 6.20\% & 7.50\% & 8.00\% & 2.50\% & 7.50\% & 8.82\% & 5.00\% & 8.57\% & 0.00\% \\
\bottomrule
\end{tabular}%
}
\end{table*}

\section{Agent Interface and Prompt Details} \label{app:agent-prompts}
This section reports the agent interfaces used in the experiments. The goal is to make the evaluation protocol reproducible without expanding the main text. All agents are evaluated under the same device setup, app reset protocol, and 50-step budget described in Section~\ref{sec:exp_setup}.

Qwen and Doubao are evaluated with their official mobile/GUI agent prompts and tool schemas. For Claude, GPT, and Gemini, we adapt the M3A agent from AndroidWorld~\cite{android_world}. The adapted interface keeps M3A's mobile observation design, action schema, Set-of-Mark grounding, and action-history mechanism. We use a single-model version rather than the original multi-agent M3A setting, so that all evaluated model families are compared under a single-model agent setup.

For all evaluated agents, we set the temperature to $0$. The action history provided at each step is a concise textual record of previous operations within the same task, including prior actions and their observed outcomes when available. It does not include past screenshots; the visual input is supplied by the current screen observation. All prompt templates are provided in Appendix~\ref{app:prompt-templates}.

\newpage
\newpage
\onecolumn
\twocolumn

\subsection{Prompt Templates} \label{app:prompt-templates}
\subsubsection{M3A Single-model GUI Prompt}
\begin{promptlisting}[M3A Single-model GUI Prompt]{promptblue}{prompt-m3a.tmp}
You are an agent who can operate an Android phone on behalf of a user. Based on user's goal/request, you may
- Answer back if the request/goal is a question (or a chat message), like user asks "What is my schedule for today?".
- Complete some tasks described in the requests/goals by performing actions (step by step) on the phone.

When given a user request, you will try to complete it step by step. At each step, you will be given the current screenshot (including the original screenshot and the same screenshot with bounding boxes and numeric indexes added to some UI elements) and a history of what you have done (in text). Based on these pieces of information and the goal, you must choose to perform one of the action in the following list (action description followed by the JSON format) by outputing the action in the correct JSON format.
- If you think the task has been completed, finish the task by using the status action with complete as goal_status: {"action_type": "status", "goal_status": "complete"}
- If you think the task is not feasible, finish by using the status action with infeasible as goal_status: {"action_type": "status", "goal_status": "infeasible"}
- Answer user's question: {"action_type": "answer", "text": "<answer_text>"}
- Click/tap on an element on the screen: {"action_type": "click", "index": <target_index>}
- Long press on an element on the screen: {"action_type": "long_press", "index": <target_index>}
- Type text into a text field: {"action_type": "input_text", "text": <text_input>, "index": <target_index>}
- Press the Enter key: {"action_type": "keyboard_enter"}
- Navigate to the home screen: {"action_type": "navigate_home"}
- Navigate back: {"action_type": "navigate_back"}
- Scroll the screen or a scrollable UI element: {"action_type": "scroll", "direction": <up, down, left, right>, "index": <optional_target_index>}
- Open an app: {"action_type": "open_app", "app_name": <name>}
- Wait for the screen to update: {"action_type": "wait"}

Guidelines:
- Usually there will be multiple ways to complete a task, pick the easiest one. If something does not work as expected, sometimes a simple retry can solve the problem, but if it does not, switch to other solutions.
- For requests that are questions, remember to use the answer action to reply to user explicitly before finish.
- If the desired state is already achieved, you can just complete the task.
- Use open_app whenever you want to open an app.
- Use input_text whenever you want to type something.
- For click, long_press and input_text, the index parameter must be visible in the screenshot and in the UI element list.
- Consider exploring the screen by using scroll actions.

The current user goal/request is: {goal}
Here is a history of what you have done so far:
{history}
The current screenshot and the same screenshot with bounding boxes and labels added are also given to you.
Here is a list of detailed information for some of the UI elements:
{ui_elements}
{additional_guidelines}
Now output an action from the above list in the correct JSON format, following the reason why you do that. Your answer should look like:
Reason: ...
Action: {"action_type":...}

Your Answer:
\end{promptlisting}

\subsubsection{Qwen-style GUI Prompt}
\begin{promptlisting}[Qwen-style GUI Prompt]{promptgreen}{prompt-qwen.tmp}
# Tools
You may call one or more functions to assist with the user query.

You are provided with function signatures within <tools></tools> XML tags:
<tools>
{"type": "function", "function": {"name": "mobile_use", "description": "Use a touchscreen to interact with a mobile device, and take screenshots. This is an interface to a mobile device with touchscreen. You can perform actions like clicking, typing, swiping, etc. Some applications may take time to start or process actions, so you may need to wait and take successive screenshots to see the results of your actions. The screen's resolution is 999x999.", "parameters": {"properties": {"action": {"description": "The action to perform. Available actions: click, long_press, swipe, type, answer, system_button, wait, terminate.", "enum": ["click", "long_press", "swipe", "type", "answer", "system_button", "wait", "terminate"], "type": "string"}, "coordinate": {"description": "Required by click, long_press, and swipe.", "type": "array"}, "coordinate2": {"description": "Required by swipe.", "type": "array"}, "text": {"description": "Required by type and answer.", "type": "string"}, "time": {"description": "Required by long_press and wait.", "type": "number"}, "button": {"description": "Required by system_button.", "enum": ["Back", "Home", "Menu", "Enter"], "type": "string"}, "status": {"description": "Required by terminate.", "enum": ["success", "failure"], "type": "string"}}, "required": ["action"], "type": "object"}}}
</tools>

For each function call, return a json object with function name and arguments within <tool_call></tool_call> XML tags:
<tool_call>
{"name": <function-name>, "arguments": <args-json-object>}
</tool_call>

# Response format
Response format for every step:
1) Thought: one concise sentence explaining the next move (no multi-step reasoning).
2) Action: a short imperative describing what to do in the UI.
3) A single <tool_call>...</tool_call> block containing only the JSON.

Rules:
- Output exactly in the order: Thought, Action, <tool_call>.
- Be brief: one sentence for Thought, one for Action.
- Do not output anything else outside those three parts.
- If finishing, use action=terminate in the tool call.

The user query: {instruction}.
Task progress (You have done the following operation on the current device): {stage2_history}.
\end{promptlisting}

\subsubsection{Doubao-style GUI Prompt}
\begin{promptlisting}[Doubao-style GUI Prompt]{promptyellow}{prompt-doubao.tmp}
You are a GUI agent. You are given a task and your action history, with screenshots. You need to perform the next action to complete the task.

## Output Format
Thought: ...
Action: ...

## Action Space
click(point='<point>x1 y1</point>')
long_press(point='<point>x1 y1</point>')
type(content='') # If you want to submit your input, use "\n" at the end of content.
scroll(point='<point>x1 y1</point>', direction='down or up or right or left')
open_app(app_name='')
drag(start_point='<point>x1 y1</point>', end_point='<point>x2 y2</point>')
press_home()
press_back()
finished(content='xxx')

## Note
- Use {language} in Thought part.
- Write a small plan and finally summarize your next action (with its target element) in one sentence in Thought part.
- The coordinate should be integer between 0~1000. Do not output float numbers.

## User Instruction
{instruction}
\end{promptlisting}

\section{Example Agent Trajectory}\label{app:case-study}

Table~\ref{tab:SR_overall_length_subtask_merged} and the analyses above report aggregate accuracy; here we walk through one solved task step by step to make the interaction pattern concrete. The task targets the Amazon simulator: \textit{Find the cheapest electronics item in the cart and report its name and price.} The agent must open the app, navigate to the cart, scroll to the bottom to see every item before comparing prices, and report the cheapest electronics item by name and price. The trajectory below uses five action types: \texttt{open\_app}, \texttt{click}, \texttt{scroll}, \texttt{answer}, and \texttt{status}.

\begin{promptlisting}[Agent Trajectory: AMAZON, Find the Cheapest Electronics Item in the Cart]{promptgreen}{case-amazon.tmp}
APP: Amazon
Goal: Find the cheapest electronics item in the cart and report its name and price.

Step 1  Reasoning : Open Amazon app first to enter the cart.
        Action    : {"action_type": "open_app", "app_name": "Amazon"}
        Screen    : App home page.

Step 2  Reasoning : Open the cart; the list is long, so only its top is visible.
        Action    : {"action_type": "click", "index": <cart_tab>}
        Screen    : Cart page, top of the list. Visible items:
                    iPhone 17 Pro Max $1369.99 ; Marshall Acton III $199.99 ;
                    MacBook Air M4 $1099.00.

Step 3  Reasoning : More items may be below; scroll down to reveal them.
        Action    : {"action_type": "scroll", "direction": "down"}
        Screen    : Two more items come into view:
                    Sony WH-1000XM5 $399.99 ; Nintendo Switch OLED $349.99.

Step 4  Reasoning : Keep scrolling to continue reading the cart.
        Action    : {"action_type": "scroll", "direction": "down"}
        Screen    : Two more items come into view:
                    Samsung Galaxy Watch7 $299.99 ; Dyson V15 Detect $649.99.

Step 5  Reasoning : The end is not yet known; keep scrolling to see more.
        Action    : {"action_type": "scroll", "direction": "down"}
        Screen    : Two more items come into view:
                    Lodge Cast Iron Cookware Set $129.99 ; KitchenAid Mixer $449.99.

Step 6  Reasoning : The end-of-list marker is now visible, so the bottom of the
                    cart is reached; stop scrolling.
        Action    : {"action_type": "scroll", "direction": "down"}
        Screen    : Last items, followed by the end of the list:
                    Nike Pegasus 41 $139.99 ; Vitamix Explorian $349.95 ;
                    Coca-Cola Zero (24-pack) $16.99.  --- Bottom of cart ---

Step 7  Reasoning : All 12 items have been seen. Among the electronics
                    (iPhone $1369.99, MacBook Air $1099.00, Sony $399.99,
                    Nintendo Switch $349.99, Galaxy Watch7 $299.99, Marshall
                    $199.99), the Marshall Acton III is the cheapest. Report its name and price.
        Action    : {"action_type": "answer", "text": "Marshall Acton III Bluetooth Home Speaker, $199.99"}

Step 8  Reasoning : The answer has been returned; the task is complete.
        Action    : {"action_type": "status", "goal_status": "complete"}

Verify Result: agent answer contains gold keys "Marshall Acton III" and "199.99"  ->  True (task solved)
\end{promptlisting}

\twocolumn

\end{document}